\documentclass[]{article}
\usepackage{arxiv}
\usepackage[english]{babel}
\usepackage{url}
\usepackage{amsmath,amssymb}
\usepackage{mathtools}
\usepackage{graphicx}
\usepackage{caption}
\usepackage{floatrow}
\newfloatcommand{capbtabbox}{table}[][\FBwidth]
\usepackage{amsthm}
\usepackage{bm}
\usepackage{subfigure}
\usepackage{epstopdf}
\usepackage{float}
\usepackage{xcolor,colortbl}
\definecolor{abstract_background}{RGB}{235,235,235}%
\definecolor{Pink}{rgb}{1,0.8,0.9}
\definecolor{LightCyan}{rgb}{0.88,1,1}
\newcolumntype{g}{>{\columncolor{Pink}}c}

\usepackage{enumitem}

\usepackage{todonotes}
\usepackage{hyperref}
\hypersetup{colorlinks=true, citecolor=blue}

\setenumerate[0]{label=\roman*)}
\setenumerate[1]{label=\arabic*.}
\setenumerate[2]{label=(\alph*)}

\newcommand{\shorttitle}{Tasks Makyth Models: Machine Learning Assisted Surrogates for Tipping Points}
\title{
Tasks makyth models: Machine learning assisted surrogates for tipping points

}
\author{
\textbf{Gianluca Fabiani\textcolor{teal}{$^{1,2}$}, Nikolaos Evangelou\textcolor{teal}{$^{2}$}, Tianqi Cui\textcolor{teal}{$^{2}$}, Juan M. Bello-Rivas\textcolor{teal}{$^{2}$,}}
\textbf{Cristina Martin-Linares\textcolor{teal}{$^{3}$,}}\\ \textbf{Constantinos Siettos\textcolor{teal}{$^{4,}$}\thanks{Corresponding author, email: \texttt{constantinos.siettos@unina.it}} \hspace{0.1cm}, Ioannis G. Kevrekidis\textcolor{teal}{$^{2,5,6,}$}\thanks{Corresponding author, email: \texttt{yannisk@jhu.edu}}} \\
{}\\
\textcolor{teal}{$^{(1)}$} Modelling Engineering Risk and Complexity, \emph{Scuola Superiore Meridionale}, Naples 80138, Italy \hspace{1cm}\\
\textcolor{teal}{$^{(2)}$}Dept. of Chemical and Biomolecular Engineering, \emph{Johns Hopkins University}, Baltimore, MD 21218, USA\\
\textcolor{teal}{$^{(3)}$}Dept. of Mechanical Engineering, \emph{Johns Hopkins University}, Baltimore, MD 21218, USA\\
\textcolor{teal}{$^{(4)}$}Dipartimento di Matematica e Applicazioni ‘‘Renato Caccioppoli", \emph{Universit\`a degli Studi di Napoli}\\ \hspace{0.38cm}\emph{Federico II}, Naples 80126, Italy\\
\textcolor{teal}{$^{(5)}$}Dept. of Applied Mathematics and Statistics, \emph{Johns Hopkins University}, Baltimore, MD 21218, USA\\
\textcolor{teal}{$^{(6)}$}School of Medicine’s Dept. of Urology, \emph{Johns Hopkins University}, Baltimore, MD 21218, USA\\}

\begin{document}

\maketitle

\begin{abstract}
\colorbox{abstract_background}{\begin{minipage}{1\linewidth}
We present a machine learning (ML)-assisted framework bridging manifold learning, neural networks, Gaussian processes, and Equation-Free multiscale modeling, 
for (a) detecting tipping points in the emergent behavior of complex systems, and (b) characterizing probabilities of rare events (here, catastrophic shifts) near them.
Our illustrative example is an event-driven, stochastic agent-based model (ABM) describing the mimetic behavior of traders in a simple financial market.
Given high-dimensional spatiotemporal data --generated by the stochastic ABM-- we construct reduced-order models for the emergent dynamics at different scales: (a) mesoscopic Integro-Partial Differential Equations (IPDEs); and (b) mean-field-type Stochastic Differential Equations (SDEs) embedded in a low-dimensional latent space, targeted to the neighborhood of the tipping point. We contrast the uses of the different models and the effort involved in learning them.
\end{minipage}}
\end{abstract}

\section{Introduction}
Complex systems are typically characterized by the occurrence of phenomena at multiple scales: from the \emph{microscopic scale}, where individual units evolve and interact among themselves, to the \emph{mesoscopic} and \emph{macroscopic scales}, where the emergent properties arise, and where system level modelling, numerical analysis, prediction, optimization and control is desired \cite{kevrekidis2003equation,karniadakis2021physics}. 
Interactions at the microscopic/individual scale may trigger abrupt changes (rare events) in the emergent dynamics, which can lead to catastrophic shifts/major irreversible changes in the dominant mesoscopic/ macroscopic spatio-temporal behavioral pattern. 
Such sudden major changes occur with higher probability near so-called \emph{tipping points} \cite{scheffer2003catastrophic,scheffer2010foreseeing,dai2012generic,van2016you,dakos2019ecosystem,armstrong2022exceeding}, which are most often associated with bifurcation points in nonlinear dynamics terminology. 
The computation of the frequency/probability of occurrence of such transitions, and the detection of the corresponding tipping points that underpin them, is of critical importance in many real-world systems ranging from financial markets, social dynamics and power networks to biological and environmental ones. 
Our need for understanding the behavior of such systems has made Agent-Based Models (ABMs) a key modeling tool for ``creating surrogate versions of real-world complex systems inside our computing machines, changing the way we do science" \cite{casti1996would,helfmann2021interacting}.

When detailed high-fidelity microscopic/atomistic simulators are available, modeling practice often involves performing extensive, brute-force temporal simulations to estimate the frequency distribution of such transitions \cite{bellorivas2016simulations, liu2015equation, voter2007introduction}. 
This direct initial approach is neither systematic nor computationally efficient for high-dimensional systems. Furthermore, it often does not provide physical insight regarding the mechanisms that drive the transitions. 
A systematic analysis of such mechanisms requires two essential tasks. First comes the discovery of an appropriate set of collective variables (observables) that can be used to describe the evolution of the emergent dynamics.  
Such coarse-scale variables may, or may not be {\em a priori} available, depending on how much physical insight we have about the problem.
Even when experience with a phenomenon suggests some physically meaningful descriptors (variables) for it, their ability to parametrize the intrinsic \hbox{(low-)} dimensional nature of the problem needs to be confirmed/tested. 
%
For this dimensionality reduction task various manifold/machine learning methods have been proposed, including diffusion maps (DMs) \cite{coifman2005geometric,nadler2006diffusion,coifman2008diffusion,singer2009detecting,lee2020coarse,galaris2022numerical}, ISOMAP \cite{balasubramanian2002isomap,bollt2007attractor,Gallos2021construction} and local linear embedding \cite{roweis2000nonlinear,papaioannou2022time}, but also autoencoders \cite{chen2018molecular,vlachas2022multiscale,floryan2022data}).\par 
Based on this initial analysis, the second task pertains to {\em the solution of the inverse problem}, i.e., the construction of appropriate reduced-order models (ROMs) for the emergent dynamics, in order to parsimoniously perform useful numerical tasks (simulations, bifurcation calculations) and --hopefully-- obtain additional physical insight.
One option is the construction of ROMs ``by paper and pencil", starting with the (first-principles) laws of the interactions between units and using tools from statistical mechanics; in this case we still need some \emph{a priori} knowledge of a ``correct`` set of macroscopic physical variables.
Clearly, such information for many real-world systems is not available, especially when we only work with data/observations. 
Furthermore, even when such information is available, the assumptions (e.g., infinite population size, homogeneous agents and network of interactions) that are made in order to obtain the closures required to bridge the gap between the micro and meso/macroscale, bias the estimations of the actual location of tipping points, as well as the statistics of the associated catastrophic shifts.\par
Another option is the data-driven identification of surrogate models in the form of ordinary, stochastic or partial differential equations via machine learning. 
Such approaches include, to name a few, sparse identification of nonlinear dynamical systems (SINDy) \cite{brunton2016discovering,schaeffer2017learning,boninsegna2018sparse}, Gaussian process regression (GPR) \cite{raissi2017machine,raissi2018numerical,lee2020coarse,chen2021solving,bertalan2019learning}, Feedforward neural networks (FNN) \cite{lee2020coarse,arbabi2020linking,galaris2022numerical,lee2023learning,dietrich2023learning,floryan2022data}, random projection neural networks (RPNN) \cite{galaris2022numerical}, recursive neural networks (RvNN) \cite{vlachas2022multiscale}, reservoir computing (RC) \cite{vlachas2020backpropagation}, neural ODEs \cite{chen2018neural,kim2021stiff}, autoencoders \cite{bertalan2019learning,li2020scalable,hasan2020identifying,floryan2022data}, DeepOnet \cite{lu2021learning}, generative adversarial networks (GANs) \cite{yang2020physics} as well as long/short-term memory (LSTM) networks \cite{vlachas2018data}.
Once again, for such machine learning based models, some insight regarding an appropriate set of collective variables is necessary. Furthermore, their approximation accuracy clearly depends very strongly on the quality of the training data, i.e., the appropriate sampling of the state and parameter space.
The latter task is far from trivial, especially around the apparent tipping points and separatrices, where the dynamics can even blow up in finite time. 
Last but not least, we confront the \emph{``curse of dimensionality”} when we require good generalization properties as we learn surrogate models from high-dimensional data.\par

If the coarse-variables are known, the Equation-free (EF) approach \cite{Makeev2002coarse,kevrekidis2003equation} offers an efficient alternative for learning ``on demand'' black-box \emph{local} coarse-grained maps for the emergent dynamics on an embedded low-dimensional subspace; this \emph{bypasses the need} to construct (global, generalizable) surrogate models, {\em focusing on the numerical task we need to perform}, and not on constructing a predictive and generalizable model.
This approach can be particularly useful when conducting numerical bifurcation analysis, or designing coarse controllers \cite{siettos2004coarse,siettos2014coarse,Siettos2012equation,patsatzis2023data}. However, even with a knowledge of good coarse-scale variables, constructing the necessary {\em lifting operator}, a crucial block in the EF framework (creating consistent initial conditions in the high-dimensional space from the low-dimensional embedded subspace where the emergent dynamics evolve) is far from trivial, especially when one tries to learn {\em distributed} mesoscopic models in the form of PDEs. 
Recently, \cite{chin2022enabling} used DMs to find a set of macroscopic variables from spatio-temporal data produced through a low-dimensional optimal velocity traffic model, the $k$-nearest neighbor algorithm to construct a lifting operator within the EF approach, and performed a numerical bifurcation analysis. 
In \cite{patsatzis2023data}, we proposed a ``next generation'' EF approach, using DMs to restrict to an embedded low-dimensional manifold, and then Geometric Harmonics, combining DMs and the Nystr\"om Extension \cite{coifman2006geometric,chiavazzo2017intrinsic,evangelou2023double} to solve the pre-image problem, i.e. to construct efficiently the lifting operator required for the EF framework. Based on such a systematic ``lift-run-embed'' loop, we proposed an approach for the data-driven design of a robust controller that --close to equilibrium points-- washes-out modeling uncertainties and numerical inaccuracies to accurately trace and stabilize the actual effective steady states.\par 

Here, based on our previous efforts on the construction of latent spaces \cite{singer2009detecting,evangelou2023double,patsatzis2023data} and ROM surrogates via machine learning \cite{lee2020coarse,galaris2022numerical,papaioannou2022time,lee2023learning,dietrich2023learning} from microscopic detailed spatio-temporal simulations, we present an integrated machine learning framework for the construction of two types of surrogate models: global as well as local. In particular, we learn (a) mesoscopic Integro Partial Differential Equations (IPDEs), and (b) --guided by the EF framework\cite{kevrekidis2003equation,kevrekidis2004equation,Siettos2012equation}-- local embedded low-dimensional mean-field Stochastic Differential Equations (SDEs), for the detection of tipping points and the construction of the probability distribution of the catastrophic transitions that occur in their neighborhood.\par 
Our main methodological point is that, given a macroscopic task, {\em it is the task itself} that determines the type of surrogate model required to perform it. In our case, we choose this task to be the identification of tipping points as well as the quantification of escape time statistics (probability of occurrence of catastrophic events) in their neighborhood.
For such tasks, a first option -common in practice- is the construction of a data-driven, ML-identified evolution equations for macroscopic fields, such as the agent density, along with the extensive (and sometimes Herculean) effort of data collection for its training. Such equations provide some physical insight for the emergent dynamics and can be used to approximate tipping points via numerical bifurcation analysis. Such an approach does not come without its problems; for example, collecting the training data and designing the sampling process in the relatively high dimensional parameter and state space, parametrized by the density of the agents, and its spatial derivatives, is not an easy task, especially in the unstable regimes where the dynamics may blow up in finite time. Furthermore, the data-driven identification of the proper macroscopic boundary conditions, as well as accounting for the problem's well-posedness, is far from obvious. 
The second option (assuming an approximate knowledge of the tipping point neighborhood in the parameter space) is to
identify a less detailed (more coarse-grained), mean-field-level, effective SDE. This offers the capability of estimating escape time statistics through either: (a) brute-force massive parallel bursts of (cheap) SDE simulations, or through (b) numerically solving a boundary-value problem for the escape times given the identified low-dimensional drift and diffusivity formulas. For such an approach, a challenging problem is that of the discovery of convenient, possibly even interpretable low-dimensional latent-spaces.\par 
Thus, in the spirit of the Wykehamist ``manners makyth man", 
 and of Rutherford Aris' ``manners makyth modellers'' \cite{aris1991manners}, in our case we argue for a ``tasks makyth models" consideration in selecting the right inverse approach for ML-assisted model selection and identification \cite{aris1991manners}.\par
Our illustrative case study is an event-driven stochastic agent-based model describing the interactions, under mimesis, of traders in a simple financial market \cite{omurtag2006modeling}. This model exhibits a tipping point, marking the onset of a financial ``bubble'' \cite{Siettos2012equation,liu2015equation}. For the particular problem, one can derive analytically a closed-form mesoscopic ``Fokker-Planck type" IPDE for the probability density of the underlying agent distribution.\par 
The structure of the paper is as follows: in Section \ref{sec:ABM}, we first describe the ABM, and the analytically derived mesoscopic Fokker-Planck-type IPDE for the emergent dynamics. In section \ref{sec:framework}, we describe the details of our framework for the ML-assisted identification of the mesoscopic IPDE, as well as an effective macroscopic mean-field SDE. Furthermore, we show how one can perform numerical bifurcation analysis based on each approach, and quantify the statistics of rare events (catastrophic shifts) through the SDE. In section \ref{sec:numerical_results}, we present our numerical results. Finally, in section \ref{sec:conclusion} we discuss capabilities and shortcomings of our ML-assisted models, and suggest future research directions.

\section{Tipping points in a financial market with mimesis}
\label{sec:ABM}
\subsection{Microscopic dynamics: the agent-based model} 
The ABM describes the interactions of a large population of, say $N$, financial traders. Each agent is described by a real-valued state variable $X_i(t) \in (-1,1)$ associated to their mood/tendency to buy or sell stocks in the financial market according to constantly updated financial news, as well as to their interactions with the other traders \cite{omurtag2006modeling}.
\begin{figure}[ht!]
    \centering
    \subfigure[]{
    \includegraphics[width=0.41\textwidth]{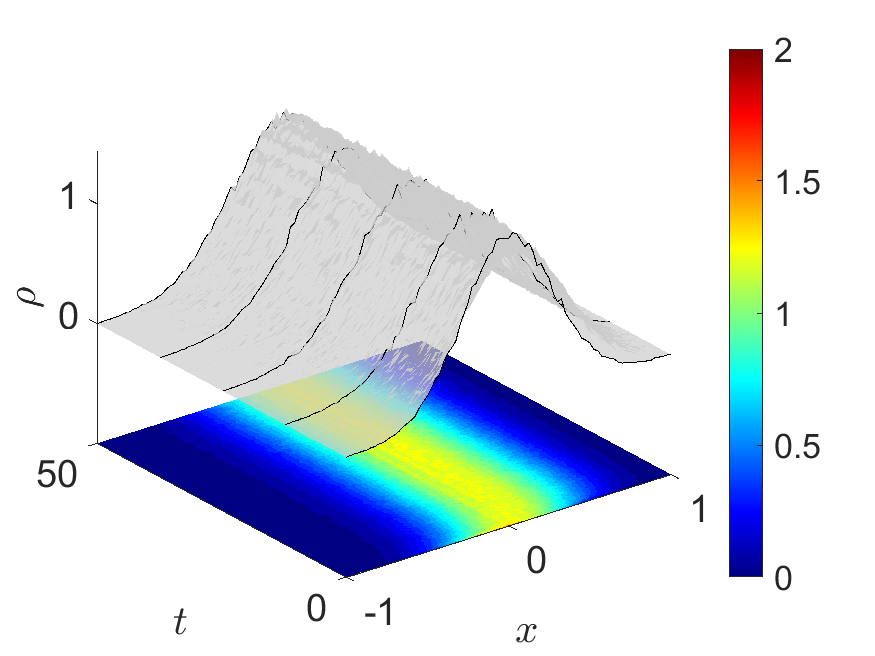}
    \label{fig:trajectories_a}
    }
    \subfigure[]{
    \includegraphics[width=0.41\textwidth]{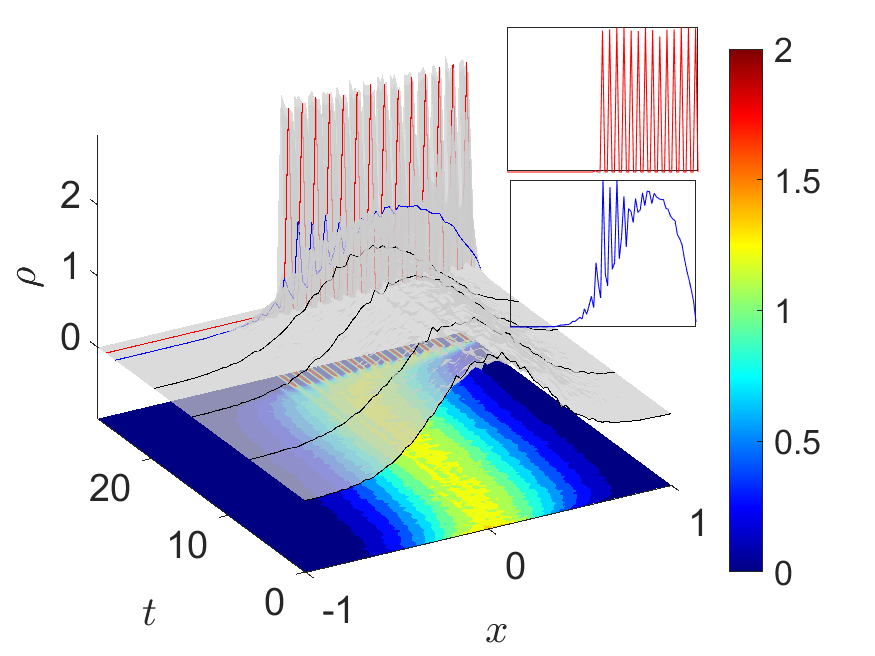}
    \label{fig:trajectories_b}
    }
    \subfigure[]{
    \includegraphics[width=0.41\textwidth]{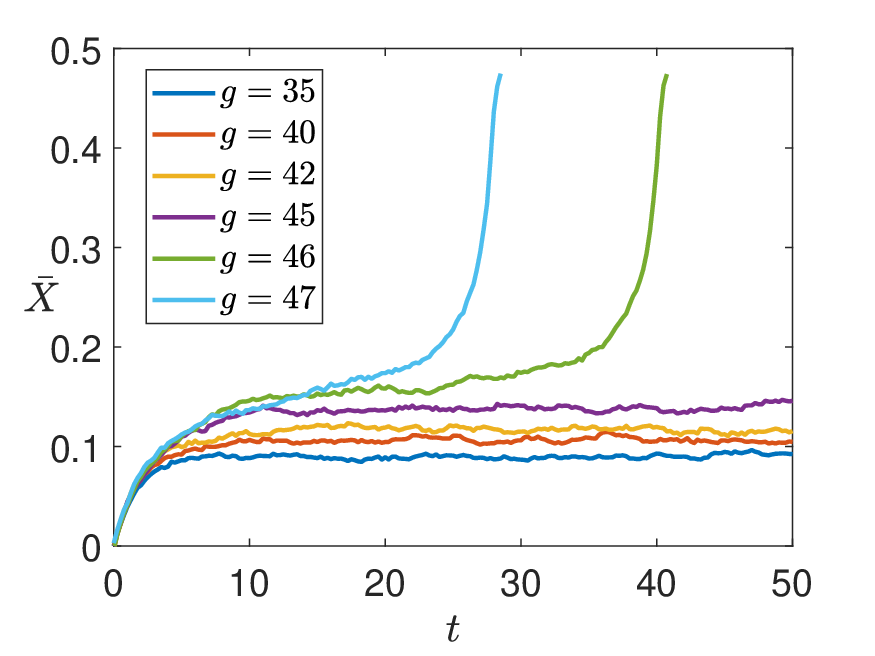}
    \label{fig:trajectories_c}
    }
    \subfigure[]{
    \includegraphics[width=0.41\textwidth]{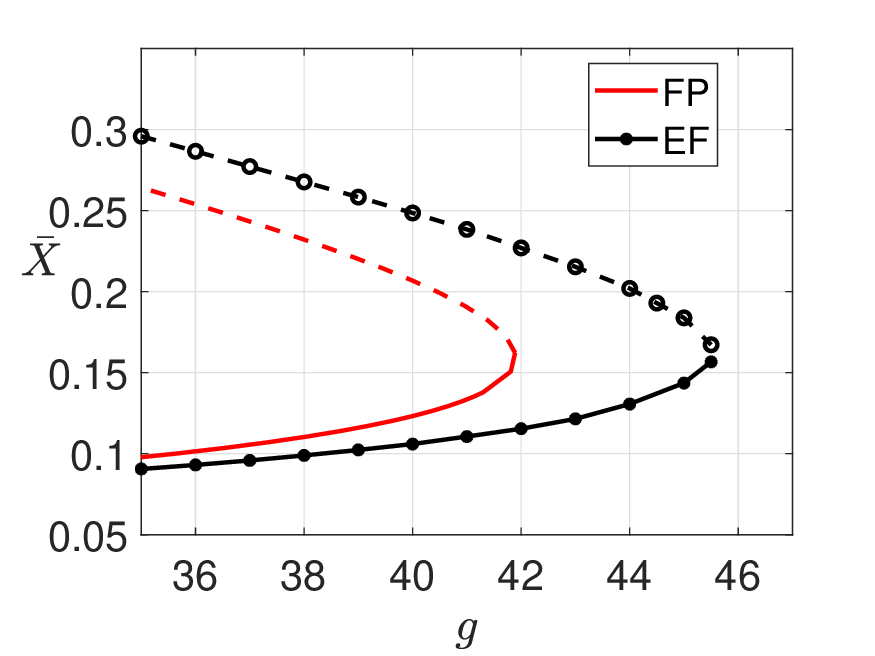}
    \label{fig:trajectories_d}
    }
    \caption{(Color version online). Agent-based model simulations. (a) Probability density function (pdf) evolution for $g=45$; (b) pdf evolution for $g=47$ (past the tipping point); Insets show the blow up of the pdf (the blue curve depicts the pdf just a few time steps before the explosion and the red curve depicts the pdf at the financial ``bubble''.  (c) Trajectories for different values of the parameter $g$; (d) Bifurcation diagram of the analytically derived Fokker-Plank (FP) vs. Equation-free (EF) computations (see \cite{siettos2014coarse,patsatzis2023data}).
    \label{fig:trajectories}}
\end{figure}
$X_i=0$ represents a neutral state, while a positive (negative) value means that the agent is more likely to buy (sell).
The $i$--th agent acts, i.e., buys or sells, only when its state $X_i$ crosses one of the decision boundaries/thresholds $X=\pm 1$. As soon as an agent $i$ buys 
or sells, 
the agent's state is forthwith reset to zero, thus assuming that its next preference does not depend on its last (buy or sell) preference.\par
The dynamics of each agent are driven by stochastic, Poisson distributed, arrivals of two types of information: {\em exogenous information} (e.g., publicly available financial news), as well as an {\em endogenous stream of information} arising from the social connections of the agents. In the absence of any incoming good news $I_i^{\!+}$ or bad news $I_i^{\!-}$, i.e., for a hypothetical isolated agent, the preference state exponentially decays to zero with a constant rate $\gamma$. Thus, each agent is governed by the following SDE:
\begin{equation}
dX_i(t)=-\gamma X_i(t)dt+dI_i^{\!+}(t)+ dI^{\!-}_i(t), \qquad |X_i|<1.
\end{equation} 
The effect of information arrivals $I^{\!\pm}_i(t)$ is represented by a series of instantaneous positive ``discrete jumps'' of size $\epsilon^{+}$, in the case of good news $I^{\!+}(t)$, and negative ``discrete jumps'' of size $\epsilon^{-}$, in the case of bad news $I^{\!-}(t)$, arriving randomly at Poisson distributed times $t_{k^{\!+}}, \, k^{\!+}= 1, 2, \dots$ and $t_{k^{\!-}}, \, k^{\!-}= 1, 2, \dots$, for good and bad news respectively. Thus, the probability $\mathbb{P}(\cdot)$ that in an interval of time $dt$ the number $N^{\!\pm}$ of arrivals is equal to $n$ is:
\begin{equation}
    \mathbb{P}(N^{\!\pm}(dt)=n)=\frac{(\nu^{\!\pm}dt)^n}{n!}e^{-\nu^{\!\pm}dt},
\end{equation}
where $\nu^{\!+}(t)$ and $\nu^{\!-}(t)$ are the average rates of arrival of the good and bad news processes in $dt$, respectively.
The agent’s state is therefore driven by two types of information terms 
\begin{equation}
I_i^{\!\pm}(t)=\sum_{k^{\!\pm}} \epsilon^{\!\pm}\delta(t-t_{k^{\!\pm}}),
\end{equation}
$\delta$ being the Dirac delta function.
Each term features two distinct arrival rates (exogenous and endogenous):
\begin{equation}
    \nu^{\!\pm}=\nu^{\!\pm}_{ex}+\nu^{\!\pm}_{en};
\end{equation}
the first is $\nu^{\!\pm}_{ex}$, the constant {\em exogenous} information arrival rate, and
the second,  $\nu^{\!\pm}_{en}$ is the {\em endogenous} arrival rate; this is shaped by the buying and selling rates averaged over all agents.
A tunable parameter $g$ embodies the strength of mimesis: the extent to which arriving information affects the willingness or apprehension of the agent to act. 
For this model, the term $\nu^{\!\pm}_{en}$ is set to be the same for all agents and is influenced by the perceived overall buying $R^{\!+}$ and selling $R^{\!-}$ rates. These are defined as the fraction of agents buying or selling per unit of time $\Delta t$:
\begin{equation}
    R^{\!\pm}=\frac{\text{number of agents buying/selling}}{\Delta t \cdot \text{total number of agents}}=\frac{1}{N\Delta t}\int_{t}^{t+\Delta t}\delta(s-T^{\!\pm}_i(s))ds,
\end{equation}
where $T^{\!\pm}_i(s)$ are the instants at which the $i$-th agent crosses the decision boundary $\pm1$ and the unit of time $\Delta t$ (the reporting horizon) is the fixed window of time at which the buying and selling rates $R^{\!\pm}$ are publicly released.
The effect of $R^{\!\pm}$ is to increase the apparent arrival rates of good/bad news according to the following expression:
\begin{equation}
\nu^{\!\pm}=\nu_{ex}^{\!\pm} + g R^{\!\pm}.
\label{eq:nu_pm}
\end{equation}
In Figures~\ref{fig:trajectories_a}-\ref{fig:trajectories_b}, we depict representative trajectories of the time evolution of the agent density distribution for $g=45,47$ and in Figure~\ref{fig:trajectories_c}, we depict the mean preference state for $N=50,000$ agents for various values of the mimesis strength $g$: $g=35,40,42,45,46,47$. The initial states of all agents were set to zero in these simulations. We see that the simulations exhibit a tipping point that arises at a parameter value $g\approx 45.5$. At the neighborhood of this tipping point, due to the inherent stochasticity of the mimetic trading process, emanate ``financial bubbles", where all agents hurry to buy assets (see Figure~\ref{fig:trajectories_c}. The ABM model also predicts financial crashes in regimes of the phase-space
where the mean value of the mesoscopic density field is negative, and the agents rush to sell (for more details see \cite{Siettos2012equation}).\par
In Figure~\ref{fig:trajectories_d}, we report the coarse-grained bifurcation diagram in terms of the first moment of the microscopic distribution, as was computed in \cite{Siettos2012equation} with the EF approach, taking as coarse-variable the distributed in space (and discretized) agent density field $\rho(x)$. The EF-based bifurcation diagram was also compared with an analytical closed-form approximation at the limit of infinite agents, in the form of an IPDE (as derived in \cite{omurtag2006modeling} where it was described as a ``Fokker-Planck-type" equation); we present it briefly in the next section. In \cite{Siettos2012equation}, we showed that the analytical ROM IPDE is not accurate enough for the expected behavior of our 50,000 agent realizations: it nontrivially underestimates the location of the tipping point with respect to $g$. In section \ref{sec:learningPDE}, we show how one can achieve a better approximation of the tipping point through data-driven black-box models in the form of IPDEs and SDEs constructed via machine learning.

\subsection{Mesoscopic dynamics: the analytically derived Fokker-Planck-type IPDE}
In order to obtain a concise description at the level of population dynamics, Omurtag and Sirovich derived a mesoscopic Fokker-Plank-type IPDE \cite{omurtag2006modeling} (a continuity equation) for the agent probability density function  $\rho(t,x)$, where the spatial variable corresponds to the preference state of the agents, i.e. $x\equiv X$. For the particular problem, at the limit of infinitely many agents and averaged along many possible trajectories, the continuity equation in terms of a probability flux $J(t,x)$ and a source $Q(t,x)$ reads:
\begin{equation}
\frac{\partial \rho(t,x)}{\partial t}=-\frac{\partial J(t,x)}{\partial x}+Q(t,x), \text{ with } J(t,x)=-\frac{1}{2} \sigma^2 \frac{\partial{\rho(t,x)}}{{\partial x}} -\mu \rho(t,x),
\end{equation}
where the source $Q(t,x)$ can be set to be $Q=(J^{\!+} + J^{\!-})\delta(x)$ to compensate for the creation/disappearance of density at the boundaries.\par
The above equation can be written as:
\begin{equation}
\frac{\partial \rho(t,x)}{\partial t}=\frac{1}{2}\sigma^2(t)\frac{\partial^2\rho(t,x)}{\partial x^2}+\frac{\partial (\mu(t,x) \rho(t,x))}{\partial x}+(J^{\!+} +J^{\!-})\delta(x).
\label{eq:FP}
\end{equation}
In the above, $\sigma^2(t)$ is the time-dependent diffusivity coefficient given by:
\begin{equation}
    \sigma^2(t)=\nu^{\!+}(\epsilon^{\!+})^2 + \nu^{\!-} (\epsilon^{\!-})^2,
    \label{eq:FP_sigma}
\end{equation}
$\mu$ is the time-dependent, space-dependent drift coefficient, given by:
\begin{equation}
    \mu(t,x)=\gamma x - \nu^{\!+}\epsilon^{\!+}-\nu^{\!-} \epsilon^{\!-}.
    \label{eq:FP_mu}
\end{equation}
$J^{\!\pm}$ denote the \emph{inflow} and \emph{outflow} through the boundaries that are restored/re-injected at the origin through a Dirac $\delta(x)$ in Eq.~\eqref{eq:FP}, in order to maintain agent conservation.
Indeed, since $\rho$ is a probability distribution, the equation has also to satisfy the normalization property:
\begin{equation}
    \int_{-1}^{1}\rho(t,x)dx=1, \quad \forall t;
\end{equation}
thus, it is convenient to consider homogeneous Dirichlet boundary conditions $\rho(\pm1,t)=0$ and the agents there instantaneously reset back to $X=0$. 
In addition, the fluxes at the boundaries $J(t,\pm 1)=J^{\!\pm}$ are given by:
\begin{equation}
J^{\!\pm}(t,x)=\mp \frac{1}{2} \sigma^2 \frac{\partial \rho(t,x)}{\partial x} \Big|_{x=\pm 1},
\end{equation}
reflecting the resetting process of agents that cross the boundaries.\par 
Besides, in order to solve/integrate the equation \eqref{eq:FP}, one has to find algebraic closures for the time-evolving diffusivity $\sigma$ and drift $\mu$ coefficients. In \cite{omurtag2006modeling}, a mean field approximation of the buying/selling rates was proposed: 
\begin{equation}
    R^{\!\pm}=\pm\nu^{\!\pm}\int_{\pm 1 \mp \epsilon^{\!\pm}}^{\pm 1} \rho(x,t) dx.
    \label{eq:Rpm_macro}
\end{equation}
Finally, based on Eqs.~\eqref{eq:nu_pm} and \eqref{eq:Rpm_macro}, one obtains:
\begin{equation}
    \nu^{\!\pm}=\frac{\nu^{\!\pm}_{ex}}{1-g\int_{\pm 1 \mp \epsilon^{\!\pm}}^{\pm 1} \rho(x,t) dx},
    \label{eq:closure_nu}
\end{equation}
from which one can retrieve at each time instance $t$, the coefficients $\mu(t)$ and $\sigma^2(t)$ of the FP model, as defined in Eqs.~\eqref{eq:FP_sigma}-\eqref{eq:FP_mu}.\par
The designation ``Fokker-Planck" notwithstanding, it is important to restate that the above approximation is an integro-differential equation with space-dependent coefficients.
\section{Methodology}
\label{sec:framework}
\begin{figure}[ht!]
    \centering
    \includegraphics[trim={0cm 5cm 0cm 0cm},clip,width=0.9\textwidth]{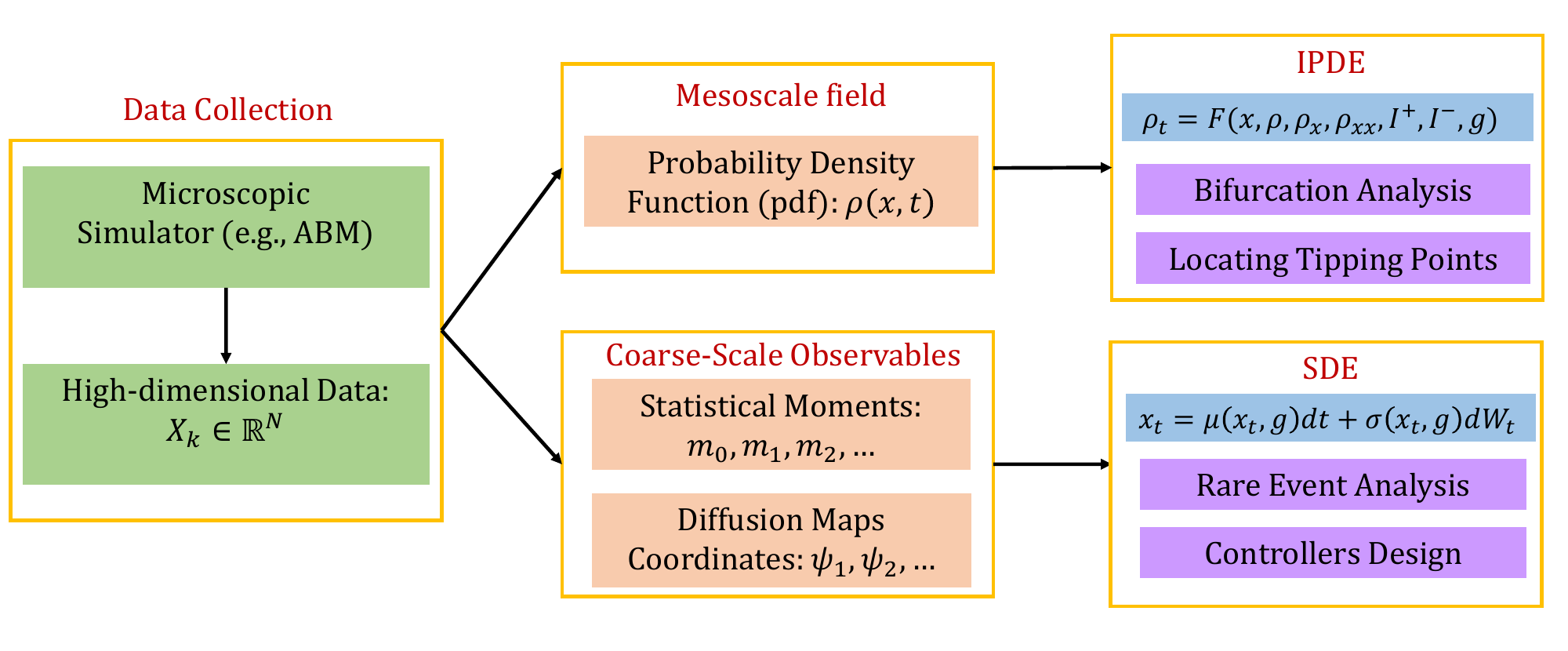}
    \caption{Schematic of the machine learning-based approach for the multiscale modelling and analysis of tipping points. At the first step, and depending on the scale of interest, we discover via Diffusion Maps latent spaces using, (a) mesoscopic fields (probability density functions (pdf) and corresponding spatial derivatives) with the aid of Automatic Relevance Determination (ARD); or (b) macroscopic mean-field quantities, such as statistical moments of the probability density function. At the second step, on the constructed latent spaces, we solve the inverse problem to identify the evolutionary laws, as IPDEs for the mesosopic field scale and mean-field SDEs for the macroscopic scale. Finally, at the third step, based on the constructed surrogate models, we perform system level analysis, such as numerical integration at a lower computational cost, numerical bifurcation analysis for the detection of tipping points, and rare event analysis for the catastrophic transitions occurring in the neighborhood of the tipping points. \label{fig:workflow}}
\end{figure}

Given high dimensional spatio-temporal trajectory data acquired through ABM simulations, the main steps of the framework are summarized as follows (see also Figure~\ref{fig:workflow}, for a schematic):
\begin{itemize}
    \item[a.] Discover low-dimensional latent spaces, on which the emergent dynamics can be described at the mesoscopic or the macroscopic scale.
    \item[b.] Identify, via machine-learning, black-box mesoscopic IPDEs, or (after furhter dimensional reduction), macroscopic mean-field SDEs.
    \item [c.] Locate tipping points  by exploiting numerical bifurcation analysis of the different surrogate models.
    \item[d.] Use the identified (NN-based) surrogate mean-field SDEs to perform rare-event analysis for the frequency of the occurrence of catastrophic transitions. This is done here in two ways: (i) performing repeated brute-force simulations around the tipping points, (ii) for this effectively 1D problem, using explicit statistical mechanics (Feynman-Kac) formulas for escape time distributions.
\end{itemize}
The above results are also validated through repeated brute-force simulations of the full agent-based model around the tipping points.
In the next sessions, we describe in detail the above elements of our approach. 

\subsection{Discovering latent spaces via Diffusion Maps}
The computational modeling of complex systems featuring a multitude of interacting agents poses a significant challenge due to the enormous number of potential states that such systems can have. Thus, a fundamental step, for the development of ROMs that are capable of effectively capturing the collective behavior of ensembles of agents is the discovery of an embedded, in the high-dimensional space, low-dimensional manifold and an appropriate set of variables that can parametrize it.\par 
Let's denote, by $\bm{X}_k \in \mathbb{R}^D$, $k=1,2,\dots$ the high-dimensional state of the ABM at time $t$. The goal is then to project/map the high-dimensional data onto lower-dimensional latent manifolds $\mathcal{M}\subset \mathbb{R}^D$, that can be defined by a set of coarse-scale variables.
The hypothesis of the existence of this manifold is related to the existence of useful ROMs and vice versa.

Here, to discover such a set of coarse-grained coordinates for the latent space, we used DMs \cite{coifman2005geometric,Coifman2006diffusion,coifman2008diffusion,singer2009detecting} (see the appendix \ref{subsec:DMs} for a brief description of the DMs algorithm).\par
For the particular ABM, we have some \emph{a priori} physical insight for the mesoscopic description. One can for example use the probability distribution function (pdf) $\rho(X)dx=\mathbb{P}(X(t)\in[x,x+dx])$ across the possible states $X_k$ in space. Thus, the continuum pdf constitutes a spatially dependent \emph{mesoscopic} field that can be modelled by a Fokker-Planck-type PDE as explained above.\par
Alternatively, one can also collect ``enough" statistical moments of the underlying distribution such as the expected value, variance, skewness, kurtosis, etc. Nevertheless, the collected statistics may not automatically provide insight into their relevance in the effective dynamics and a further feature selection/sensitivity analysis may be needed.\par 
Focusing on a reduced set of coarse-scale variable is particularly relevant when there exists a significant separation of time scales in the system's dynamics. By selecting only a few dominant statistics, one can effectively summarize the behavior of the system at a coarser level.\par 
The choice of the scale and details of coarse-grained description, leads to different modelling approaches. For example, focusing at the \emph{mesoscale} for the population density dynamics, we aim at constructing a Fokker-Planck-type IPDE. At an even coarser scale, e.g., for the first moment of the distribution, and taking into account the underlying stochasticity, a natural first choice is the construction of a mean-field macroscopic SDE. Here, we construct surrogate models via machine learning at both these distinct coarse-grained scales.

\subsection{Learning mesoscopic IPDEs via neural networks}
\begin{figure}[ht!]
    \centering
    \subfigure[]{
\includegraphics[width=0.65\textwidth]
{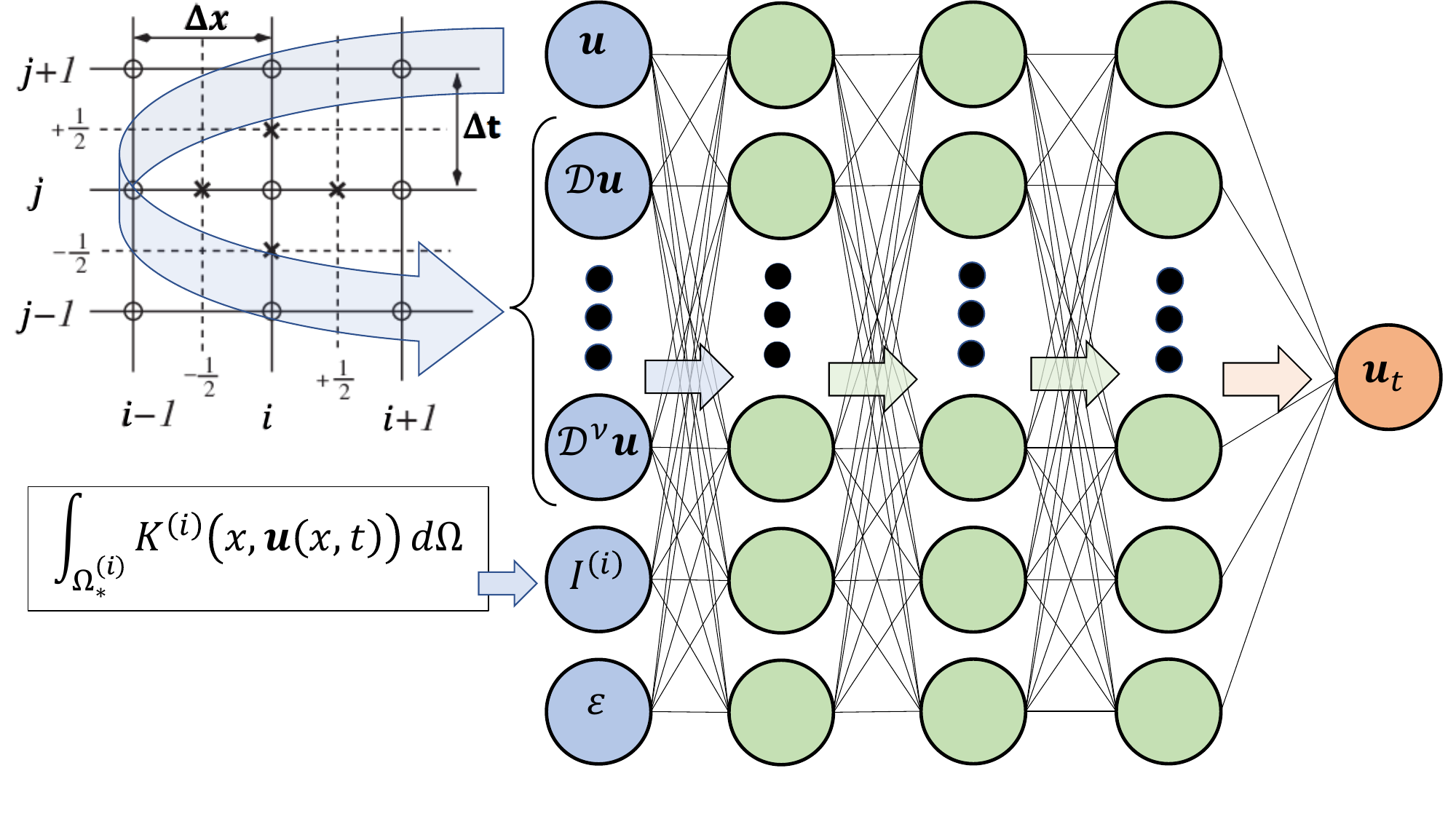}
\label{fig:schematic_NN_a}}
\subfigure[]{
\includegraphics[trim={0.5cm 1.9cm 0.2cm 1.8cm},clip,width=0.81 \textwidth]{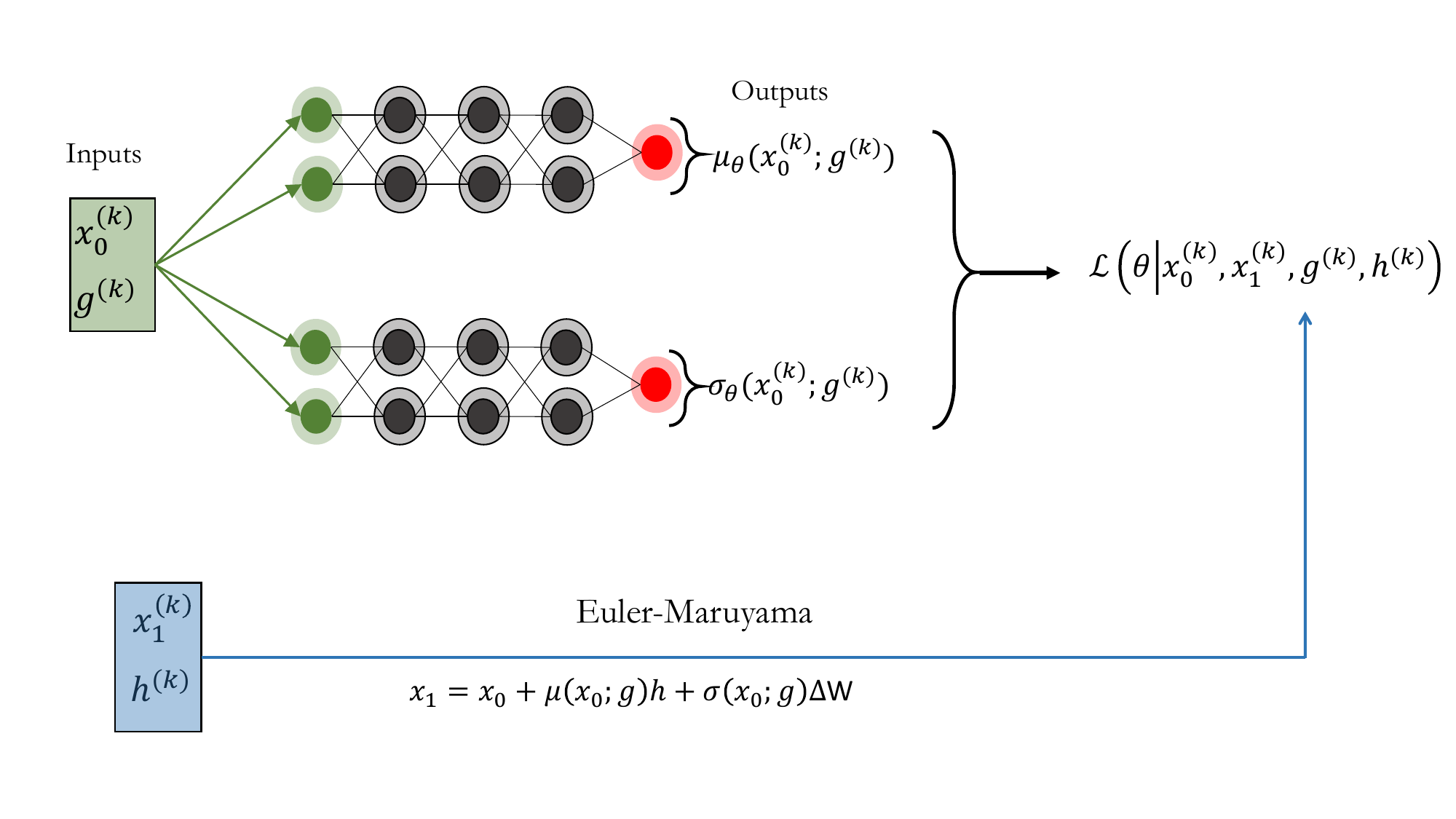}
\label{fig:schematic_NN_b}
}
    \caption{(a) Schematic of a Feedforward Neural Network (FNN), where the input is constructed by convolution operations, i.e. a combination of (a) sliding Finite Difference (FD) stencils, and, (b) integral operators, for learning mesoscopic models in the form of IPDEs (eq.~\eqref{eq:IPDEs}); the inputs to the RHS of the IPDE are the features in eq.~\eqref{eq:full_feature_space}.
    (b) A schematic of the neural network architecture, inspired by numerical stochastic integrators, used to construct macroscopic models in the form of mean-field SDEs.\label{fig:schematic_NN}}
\end{figure}

As we have discussed in the introduction, the identification of evolution operators of spatio-temporal dynamics using machine learning tools, including deep learning and Gaussian processes, represents a well-established field of research. 
The main assumption here is that the emergent dynamics of the complex system under study on a domain $\Omega\times[t_0, t_{end}] \subseteq \mathbb{R}^d \times \mathbb{R}$ can be modelled by a system of say $m$ IPDEs in the form of:
\begin{equation}\label{eq:IPDEs}
\begin{aligned}
    \frac{\partial u^{(i)}(\bm{x},t)}{\partial t}\equiv u_t^{(i)}=
     F^{(i)}(\bm{x},\bm{u}(\bm{x},t),\mathcal{D}\bm{u}(\bm{x},t),\mathcal{D}^{\bm{2}}\bm{u}(\bm{x},t),\dots,\mathcal{D}^{\bm{\nu}}\bm{u}(\bm{x},t),I_1^{(i)}(\bm{u}),I_2^{(i)}(\bm{u}),\dots,\bm{\varepsilon}),\\
     (\bm{x},t)\in \Omega\times[t_0,t_{end}], \qquad i=1,2,\dots,m,\\
    \end{aligned}
\end{equation}
where $F^{(i)},$ $i=1,2,\dots m$ are $m$ non-linear integro-differential operators; $\bm{u}(\bm{x},t)=[u^{(1)}(\bm{x},t),\dots,u^{(m)}(\bm{x},t)]$ is the vector containing the spatio-temporal fields, $\mathcal{D}^{\bm{\nu}}\bm{u}(\bm{x},t)$ is the generic multi-index $\bm{\nu}$-th order spatial derivative at time $t$ i.e.,:
\begin{equation*}
    \mathcal{D}^{\bm{\nu}}\bm{u}(\bm{x},t):=\left \{ \frac{\partial^{|\bm{\nu}|}\bm{u}(\bm{x},t)}{\partial x_1^{\nu_1}\cdots\partial x_d^{\nu_d}} , |\bm{\nu}|=\nu_1+\nu_2+\dots+\nu_d,\, \nu_1,\dots,\nu_d\ge0 \right \},
\end{equation*}
$I_1^{(i)}, I_2^{(i)},\dots$ are a collection of integral features on subdomains $\Omega_{1}^{(i)},\Omega_{2}^{(i)},\dots \subseteq\Omega$:
\begin{equation}
    I_j^{(i)}(\bm{u})=\int_{\Omega^{(i)}_j} K_j^{(i)}(\bm{x},\bm{u}(\bm{x},t))d\Omega, \qquad j=1,2,\dots;
\end{equation}
$K_j^{(i)}:\mathbb{R}^{d} \times \mathbb{R}^m\mapsto \mathbb{R}^d$ are nonlinear maps and $\bm{\varepsilon} \in \mathbb{R}^p$ denotes the (bifurcation) parameters of the system.
The right-hand-side of the $i$-th IPDE depends on say, a number of $\gamma^{(i)}$ variables and on bifurcation parameters from the set of features:
\begin{equation}
   \mathcal{S}^{(i)}=\{\bm{x},\bm{u}(\bm{x},t),\mathcal{D}\bm{u}(\bm{x},t),\mathcal{D}^{\bm{2}}\bm{u}(\bm{x},t),\dots,\mathcal{D}^{\bm{\nu}}\bm{u}(\bm{x},t),I_1^{(i)},I_2^{(i)},\dots,\bm{\varepsilon}\}.
   \label{eq:full_feature_space}
\end{equation}
At each spatial point $\bm{x}_q,q=1,2,\dots,M$ and time instant $t_s,s=1,2,\dots,N$, a single sample point (an observation) in the set $\mathcal{S}^{(i)}$ for the $i$-th IPDE can be described by a vector $Z^{(i)}_j\equiv Z^{(i)}_{(q,s)}\in \mathbb{R}^{\gamma^{(i)}}$, with $j=q+(s-1)M$.
Here, we assume that such mesoscopic IPDEs in principle exist, but they are not available in closed-form. Henceforth, we aim to learn the macroscopic laws by employing a Feedforward Neural Network (FNN), in which the effective input layer is constructed by a finite stencil (sliding over the computational domain), mimicking convolutional operations where the applied ``filter" involves values of our field variable(s) $u^{(i)}$ on the stencil, and returns features $Z_j^{(i)}\in\mathcal{S}^{(i)}$ of these variables at the stencil center-point, i.e. spatial derivatives as well as (local or global) integrals (see  Figure~\ref{fig:schematic_NN_a} for a schematic). 
The FNN is fed with the sample points $Z^{(i)}_j\in \mathbb{R}^{\gamma^{(i)}}$ and its output is an approximation of the time derivative $\bm{u}_t$. In Appendix \ref{subsec:FNN}-\ref{subsec:RPNN} we describe two different approaches (gradient descent and random projections) to train such a FNN. For alternative operator-learning approximation methods see e.g. \cite{li2020fourier} and \cite{lu2021learning}.

\subsubsection{Feature selection with Automatic Relevance Determination}
\label{subsec:ARD}
Training the FNN with the ``full'' set of inputs $\mathcal{S}^{(i)} \subseteq \mathbb{R}^{\gamma^{(i)}}$, described in~\eqref{eq:full_feature_space}, consisting of all local mean field values as well as all their coarse-scale spatial derivatives (up to some order $\nu$) is prohibitive due to the \emph{curse of dimensionality}. Therefore, one important task for the training of the FNN is to extract a few ``relevant''/dominant variable combinations. 
Towards this aim, we used Automatic Relevance Determination (ARD) in the framework of Gaussian processes regression (GPR) \cite{barber2012bayesian}. The approach assumes that the collection of all observations $\bm{Z}^{(i)}=(Z^{(i)}_1, Z^{(i)}_2,\dots, Z^{(i)}_{MN})$, of the features $z_l \in \mathcal{S}^{(i)}$, are a set of random variables whose finite collections have a multivariate Gaussian distribution with an unknown mean (usually set to zero) and an unknown covariance matrix $K$. This covariance matrix is commonly formulated by a Euclidean distance-based
kernel function $k$ in the input space, whose hyperparameters are optimized based on the training data. Here, we employ a radial basis kernel function (RBF), which is the default kernel function in Gaussian process regression, with ARD:
\begin{equation}   K_{jh}^{(i)}=k(Z^{(i)}_j,Z^{(i)}_h,\bm{\theta}^{(i)})=\theta_0^{(i)} exp\biggl( -\frac{1}{2}\sum_{l=1}^{\gamma^{(i)}} \frac{z_{l,j}-z_{l,h}}{\theta_l^{(i)}}\biggr);
    \label{eq:gaussian_processes}
\end{equation}
$\bm{\theta}^{(i)}=[\theta_0^{(i)},\theta_1^{(i)},\dots,\theta_{\gamma^{(i)}}^{(i)}]$ are  $(\gamma^{(i)}+1)$--dimensional vectors of hyper-parameters. The optimal hyperparameter set $\bm{\tilde{\theta}}^{(i)}$ can be obtained by minimizing a negative log marginal likelihood over the training data set $(\bm{Z}^{(i)},\bm{Y}^{(i)})$, with inputs the observation $\bm{Z}^{(i)}$ of the set $\mathcal{S}^{(i)}$ and corresponding desired output given by the observation $\bm{Y}^{(i)}$ of the time derivative $u^{(i)}_t$:
\begin{equation}
    \bm{\tilde{\theta}}^{(i)}=arg\min_{\bm{\theta}^{(i)}} -\log{p(\bm{Y}^{(i)}|\bm{Z}^{(i)},\bm{\theta}^{(i)})}.
\end{equation}
As can be seen in Eq.~\eqref{eq:gaussian_processes}, a large value of $\theta_l$ nullifies the difference between target function values along the $l$-th dimension, allowing us to designate the corresponding $z_l$ feature as ``insignificant''.
Practically, in order 
to build a reduced input data domain, we define
the normalized effective \emph{relevance weights} $W_r^{(i)}(\cdot)$ of each feature input $z_l \in \mathcal{S}^{(i)}$, by taking:
\begin{equation}
    \bar{W}_r^{(i)}(z_l) = exp(-\tilde{\theta}^{(i)}_l), \qquad W_r^{(i)}(z_l)=\frac{\bar{W}_r^{(i)}(z_l)}{\sum_l \bar{W}^{(i)}_r(z_l)}.
\end{equation}
Thus, we define a small tolerance $tol$ in order to disregard the components such that $W_r^{(i)}(z_l)<tol$.
The remaining selected features ($W_r^{(i)}(z_l)\ge tol$) can still successfully (for all practical purposes) parametrize the approximation of the right-hand-side of the underlying IPDE.


\subsection{Macroscopic mean-field SDEs via neural networks}
Here, we present our approach for the construction of embedded surrogate models in the form of mean-field SDEs. Under the assumption that we are close (in phase- and parameter space) to a previously located tipping point, we can reasonably assume that the effective dimensionality of the dynamics can be reduced to the corresponding normal form. We already have some qualitative insight on the type of the tipping point, based for example on the numerical bifurcation calculations that located it (e.g. from the analytical-derived Fokker-Planck-type PDE, from our surrogate IPDE \cite{kalia2021learning}, or from the EF analysis \cite{Siettos2012equation}). For this particular problem, we have found that the tipping point corresponds to a saddle-node bifurcation.\par

Given the nature of the bifurcation (and the single variable corresponding normal form) we identify a one-dimensional SDE, driven by a Wiener process, from data. We note that learning higher-order such SDEs, or SDEs based on the more general L\'evy process \cite{fang2022end}, and the Ornstein–Uhlenbeck process \cite{karatzas1991brownian} is straightforward. \par
For a diffusion process with drift, say $X_t=\{x_t, t>0\}$, the drift, $\mu(x_t)$ and diffusivity $\sigma^2(x_t)$ coefficients over an infinitesimally small-time interval $dt$, are given by:
\begin{align}
    \mu(x_t)=\lim_{\delta t \rightarrow 0} \frac{1}{\delta t}\mathbb{E}(\delta x_t|X=x_t), \quad  \sigma^2(x_t)=\lim_{\delta t \rightarrow 0} \frac{1}{\delta t}\mathbb{E}{(\delta x_t)}^2|X=x_t),
\end{align}
where, $\delta x_t =x_{t+\delta t}-x_t$.\par 
The 1D SDE driven by a Wiener process $W_t$ reads:
\begin{equation}
    d x_t = \mu(x_t;g) dt + \sigma(x_t;g) d W_t. \label{eq:SDE}
\end{equation}

Here, for simplicity, we assume that the one-dimensional parameter $g$, introduced in Eq.~\eqref{eq:nu_pm}, enters into the dynamics, via the drift and diffusivity coefficients. 
Our goal is to identify the functional form of the drift $\mu(x,g)$ and the diffusivity $\sigma(x,g)$ given noisy data close to the tipping point via machine learning. For the training, the data might be collected from either long-time trajectories or short bursts initialized at scattered snapshots, as in the EF framework. These trajectories form our data set of input-output pairs of discrete-time maps. 
A data point in the collected data set can be written as $(x_0^{(k)}, h^{(k)}, x_1^{(k)} g^{(k)})$, where $x_0^{(k)}$ and $x_1^{(k)}$ measures two consecutive states at $t_0^{(k)}$ and $t_1^{(k)}$ with (small enough) time step $h^{(k)} = t_1^{(k)} - t_0^{(k)}$ and $g^{(k)}$ is the parameter value for this pair. Based on the above formulation, one goes from $x_0^{(k)}$ to $x_1^{(k)}$ by:
\begin{equation}
    x_1^{(k)} = x_0^{(k)} + \int_{t_0^{(k)}}^{t_1^{(k)}} \mu(x_t;g) dt + \int_{t_0^{(k)}}^{t_1^{(k)}} \sigma(x_t;g) d W_t.
    \label{eq:SDE_int}
\end{equation}
Here, for the numerical integration of the above equation to get stochastic realizations of $x_1^{(k)}$, we assume that the Euler-Maruyama numerical scheme   can be used, reading:
\begin{equation}
    x_1^{(k)} \approx x_0^{(k)} + h^{(k)} \mu(x_0^{(k)};g^{(k)}) + \sigma(x_0^{(k)};g^{(k)}) \Delta W^{(k)}, \label{eq:SDE_Euler}
\end{equation}
where $\Delta W^{(k)} = W_{t_1^{(k)}} - W_{t_0^{(k)}} \in \mathbb{R}$ is a one-dimensional random variable, normally distributed with expected value zero and variance $h^{(k)}$.\par

Considering the point $x_1^{(k)}$ as a realization of a random variable $X_1$, conditioned on $x_0^{(k)}$ and $h^{(k)}$, drawn by a Gaussian distribution of the form:
\begin{equation}
    p_{X_1}(x_1^{(k)})=\mathbb{P}\left(X_1=x_1^{(k)} \mid X_0 = x_0^{(k)}, h^{(k)}\right) \sim \mathcal{N} \bigg(x_0^{(k)} + h^{(k)} \mu(x_0^{(k)};g^{(k)}), h^{(k)} \sigma(x_0^{(k)};g^{(k)})^2 \bigg), \label{eq:cond_prob}
\end{equation}

we approximate the drift $\mu(x_0^{(k)};g^{(k)})$ and diffusivity $\sigma(x_0^{(k)};g^{(k)})$ functions by simultaneously training two neural networks, denoted as $\mu_{\theta}$ and $\sigma_{\theta}$, respectively. This training process involves minimizing the loss function: 
\begin{equation}
    \label{eq:loss_sde_nn}
    \mathcal{L}(\theta|x_0^{(k)},x_1^{(k)},h^{(k)}) :=\sum_{k} \frac{(x_1^{(k)} - x_0^{(k)} - h^{(k)}\mu_{\theta}(x_0^{(k)};g^{(k)}))^2}{h^{(k)}\sigma_{\theta}(x_0^{(k)};g^{(k)})^2}  + \text{log}|h^{(k)}\sigma_{\theta}(x_0^{(k)};g^{(k)})^2|.
\end{equation}
which is derived in order to maximize the log-likelihood 
of the data and where $\theta$ denotes the trainable parameters (e.g., weights and biases of the neural networks $\mu_\theta$ and $\sigma_\theta$). A schematic of the Neural Network, based on Euler-Maruyama, is shown in Figure~\ref{fig:schematic_NN_b}.

\subsection{Locating tipping points via our surrogate models}
\label{sec:locate}
In order to locate the tipping point, based on either the mesoscopic IPDE or the embedded mean-field 1D SDE model, we construct the corresponding bifurcation diagram in its neighborhood, using pseudo-arc-length continuation as implemented in numerical bifurcation packages.

For the identified SDE, we used its deterministic part, i.e., the drift term, to perform continuation. 
The required Jacobian of the activation functions of the neural network is computed by symbolic differentiation. Note that this is just a validation step (we already know the location and nature of the tipping point).

\subsection{Rare-event analysis of catastrophic shifts} \label{sec:MFPT}
Given a sample space $\Omega$, an index set of times $T$ and a state space $S$, the \emph{first passage time}, also known as \emph{mean exit time} or \emph{mean escape time}, of a stochastic process $x_t: \Omega \times T \mapsto S$ \textit{on a measurable subset} $A \subseteq S$ is a random variable which can be defined as
\begin{equation}
    \tau(\omega) := \inf \{t \in T \mid x_t(\omega)\in S \setminus A\}, \label{eq:FPT}
\end{equation}
where $\omega$ is a sample out of the space $\Omega$. One can define the mean escape time from $A$, which works as the expectation of $\tau(\omega)$:
\begin{equation}
    \langle \tau \rangle := \mathbb{E}_{\omega} [\tau(\omega)]. \label{eq:MFPT}
\end{equation}
For a $n$-dimensional stochastic process, as it is the ABM under study, $S$ is typically set to be $\mathbb{R}^n$, and $A$ is usually a bounded subset of $\mathbb{R}^n$. In the case of our local 1D SDE model (see Section \ref{sec:learningSDE}), the subset $A$ reduces to an open interval $(a, b)$, with the initial condition of this stochastic process $x_0$ also chosen in this interval.\par 
We discuss two ways for estimating the occurrence of those rare-events. The first involves direct computational ``cheap'' temporal simulations of the 1D SDE, where one gets an empirical probability distribution; the second is a closed-form expression, based on statistical mechanics, for the mean escape time (assuming an exponential distribution of escape times).

\paragraph{Computation of escape times based on temporal simulations of the 1D SDE}
We perform numerical integration of multiple stochastic trajectories of the SDE to obtain an estimation of the desired mean escape time. The algorithm for estimating the escape times of a one-dimensional stochastic process with initial condition $x_0$ on the interval $(a, b)$ can be described as follows:
\begin{enumerate}    
    \item Given a fixed time step $h > 0$, we perform $i+1$ numerical integration steps of the SDE until a stopping condition, i.e., $x_t$ exits, for the first time $t_{i+1}$, the interval at $a$ or $b$.
    \item Record the time $\tau = t_{i} $, which corresponds to a realization of the escape time.
    \item Repeat the above steps 1-2 for $K$ iterations, and each time collect the observed escape time.
    \item Compute the statistical mean value of the collected escape times that corresponds to an approximation of the mean escape time.
\end{enumerate}
In our case, the initial condition $x_0$ was set equal to the stable steady state, and the termination condition $a$ or $b$ at which we consider the dynamics \textit{escaped/exploded}.

\paragraph{Statistical mechanics-based computation of escape times}
\label{sec:quad}
The second approach for estimating escape times involves numerical computation of the escape times based on the known closed-form expression for the 1D SDE. Here we report the expressions for the constant diffusivity case, since the identified diffusivities in our case appear not to be state (or parameter) dependent, see Section \ref{sec:learningSDE}.
The effective potential $U_e$ for a 1D SDE with constant diffusivity is given by:
\begin{equation}
    \label{eq:effective_potential}
    U_e(x) = \int_{x_0}^x -\frac{\mu(x)}{\sigma^2/2} dx,
\end{equation}
where $x_0$ is an arbitrarily selected reference point in which the potential is taken to be zero, $\mu(x)$ is the drift, and $\sigma$ is the diffusivity. Equation \eqref{eq:effective_potential} is derived from the  steady-state solution of the Fokker-Planck
equation, see Appendix C in \cite{frewen2009exploration} (for a detailed derivation see also \cite{risken1996fokker}).
It has been pointed out in \cite{bellorivas2016simulations} that, there is a closed-form expression of escape times for one-dimensional systems with constant diffusivity \eqref{eq:Juan_SDE}.
\begin{equation}
    d x_t = - \nabla U_e(x_t) dt + \sqrt{\frac{2}{\beta}} d W_t, x_0 \in (a, b), \label{eq:Juan_SDE}
\end{equation}
where $U_e(x_t)$ is the effective potential of the system, 
$(a, b)$ are the left and right boundaries of the interval of interest, and $\beta$ is the inverse temperature, inspired from thermodynamics.
Compared with \eqref{eq:SDE}, we can transform our learned SDE form into \eqref{eq:Juan_SDE} by setting
\begin{equation}
    \mu(x_t) = -\nabla U_e(x_t) = - \frac{dU_e(x_t)}{d x_t}; \sigma = \sqrt{\frac{2}{\beta}}. \label{eq:equi_SDE}
\end{equation}
Notice that one can also consider the drift term $\mu(x_t)$ as the force term in an energy field. The mean escape time for systems like \eqref{eq:Juan_SDE} can be written as a univariate integral:
\begin{equation}
    \langle \tau \rangle = \int_a^b \rho(x; x_0) dx, \label{eq:Juan_MFPT}
\end{equation}
where $\rho(x; x_0)$ is the occupation density of particles at $x = x_0$ which solves the boundary value problem (see Section~\ref{sec:quad}),
\begin{equation*}
  \left\{
    \begin{aligned}
      -&\frac{d}{d x}
      \left(
      \beta^{-1} \rho^\prime(x) + \rho(x) U^\prime(x)
      \right)
      =
         \delta(x - x_0),
      & \text{in $(a, b)$,} \\
      &\rho(a) = \rho(b) = 0.
    \end{aligned}
  \right.
\end{equation*}
The occupation density $\rho$ can be computed as the solution of the above boundary value problem:
\begin{equation}
    \rho(x; x_0) = \beta \int_a^x (G(x_0) - H(z - x_0)) e^{\beta (U(z) - U(x))} dz, \label{eq:Juan_density}
\end{equation}
where $H(\cdot)$ is the Heaviside step function and $G(\cdot)$ is defined as
\begin{equation}
    G(x_0) = \frac{\int_{x_0}^b e^{\beta U(\eta)} d \eta}{\int_a^b e^{\beta U(\eta)} d \eta}. \label{eq:Juan_G}
\end{equation}
The above analysis does not depend on the choice of potential reference point. 
Consider another representation of the potential $U_e^{\prime}(\cdot)$ that actually has
\begin{equation}
    U_e^{\prime}(\eta) - U_e(\eta) = C \label{eq:other_U}
\end{equation}
for $\forall \eta \in \mathbb{R}$, where $C$ is a constant. One can substitute \eqref{eq:other_U} in \eqref{eq:Juan_G} to get
\begin{equation}
    G'(x_0) = \frac{\int_{x_0}^b e^{\beta U_e^{\prime}(\eta)} d \eta}{\int_a^b e^{\beta U_e^{\prime}(\eta)} d \eta} =\frac{\int_{x_0}^b e^{\beta (U_e(\eta) + C)} d \eta}{\int_a^b e^{\beta (U_e(\eta) + C)} d \eta} =\frac{e^{\beta C} \int_{x_0}^b e^{\beta U_e(\eta)} d \eta}{e^{\beta C} \int_a^b e^{\beta U_e^{\prime}(\eta)} d \eta} = G(x_0);
\end{equation}
\noindent this implies that the function $G(\cdot)$ is independent of the reference point. As the exponential term in \eqref{eq:Juan_density} is a function of the difference of potential values, the choice of the reference point will not affect the value of the occupation density $\rho(\cdot)$, nor the values of escape times $\langle \tau \rangle$.

In practice, the integration can be performed numerically using a quadrature scheme.
However, quadrature schemes are susceptible to numerical errors under some circumstances.
For example, if the diffusivity is small, $\sigma \ll 1$ (and thus $\beta$ is large), this leads to exponential values that might be numerically intractable and impede the utilization of the method.

To address this issue, we transform the SDE~\eqref{eq:SDE} by $x \mapsto y(x) = x / \sigma$ and apply It\^o's lemma to obtain the SDE,
\begin{equation}
    d y_t = \sigma^{-1} \mu(\sigma y_t;g) dt + dW_t. \label{eq:SDE_transform}
\end{equation}
where the diffusivity in this case is one.

This transformation allows us to compute the escape times in terms of the transformed variables $y_t$ and using $\beta = 2$, which alleviates the numerical issues. The escape times for the transformed variable $y_t$ and the original variables 
are the same. This allows us to transform the SDEs identified from the trained neural networks to compute the escape times using~\eqref{eq:Juan_MFPT}.
It is worth mentioning that the applied transformation, even though it circumvents the numerical issues arising because of the diffusivity in cases where the potential is large (and therefore the exponent of the potential is also large), would still be susceptible to numerical issues in other regimes.

\section{Numerical results}
\label{sec:numerical_results}
\subsection{Tipping points via the learned mesoscopic IPDE}
\label{sec:learningPDE}
\subsubsection{Training and test data sets}
ABM simulations were performed using $N=50,000$ agents, with $\nu^{\!+}_{ex}=\nu^{\!-}_{ex}=20,$ $ \gamma=1$, $\epsilon^{\!-}=-0.072$,  $\epsilon^{\!+}=0.075$. The mimetic strength  $g$ is our bifurcation parameter. For the data set, we selected 41 equally-spaced points in the range $g \in [30,50]$ and for each of the values of $g$, we randomly generated 1000 different initial profiles $\rho_0$, as Gaussian distributions $\rho_0 \sim \mathcal{N}(\tilde{m}_0,\tilde{s}^2_0)$ with varying mean $\tilde{m}_0$ and variance $\tilde{s}^2_0$. The initial $\tilde{m}_0$ and $\tilde{s}_0$ were uniformly randomly sampled as $\tilde{m}_0 \sim \mathcal{U}([-0.3,0.3])$, $
\tilde{s}_0 \sim \mathcal{U}([0.3,0.5])$. 
The initial state of the agents is sampled from the initial distribution $\rho_0$, creating a consistent microscopic realization.
At each time step, as the agents dynamically evolve, to estimate the corresponding coarse-grained density profile, we used $81$ equally-spaced bins, with equally-spaced centers $x_i \in [-1,1]$.

Since we are dealing with a stochastic model, in order to generate smooth enough profiles for the spatial derivatives, for each fixed initial condition, we ran 100 random stochastic realizations, and we averaged along the generated copies of the density field. Then to further smooth out the computed densities, we applied a weighted moving average smoothing $\rho_i$  as $\rho_i^{\star}=\frac{2\rho_i+\rho_{i-1}+\rho_{i+1}}{4}.$ ($i$ denotes spatial mesh points).

The ABM simulations were run for a time interval $t \in [0,15]$ and we collected points with a time step of $\Delta t=0.25$. When the mean value $\bar{X}$ crossed, $\pm 0.4$ we stopped the simulations, because the pdf profile blows up, very fast after that. Furthermore, we also ignored the first two time points, to exclude the initial ``healing'' transients due to the way we initialize (see the discussion for such healing periods in \cite{kevrekidis2003equation}). We thus end up with a data set consisting of
$40$ (values of $g$) $\times 100$ (initial conditions) $\times 58$ (maximum time points ignoring the first 2 steps of the transient) $\times 81$ (space points) $\approx$ $15\times 10^6$ data points.
Since the amount of data is practically too large, for the training set we have randomly downsampled to $10^6$ data and used the remaining data as our test set.

\subsubsection{Feature selection for the mesoscopic IPDE model}
For dealing with the ``curse of dimensionality'' in training the FNN, learning our IPDE model, we used ARD as implemented in \texttt{Matlab} by the function \textit{fitrgp} for feature selection. Here, we \emph{a priori} selected as candidate features, the space $x$ \textit{per se}, the field $\rho$, the first $\rho_x$, second $\rho_{xx}$ and third $\rho_{xxx}$ spatial derivatives estimated with central finite differences, as well as $I^{\!\pm}$ defined as the integrals of $\rho$ in a small region close to the boundaries:
\begin{equation}
    I^{\!\pm}(t)=\pm\int_{\pm 1 \mp \epsilon}^{\pm 1}\rho(x,t)dx,
\end{equation}
where $\epsilon=0.05$ corresponds to the size of the last two bins of the grid. We note that the latter two candidate mesoscopic variables $I^{\!\pm}$ as defined above are related to the buying and selling rates (see Eq.\ref{eq:Rpm_macro}), yet they do not depend on the frequencies $v^{\!\pm}$ as the true buying and selling rates do. These ``internal''/hidden variables $v^{\!\pm}$ are considered unknown. We also consider unknown the ``quantum'' jump sizes $\epsilon^{\!\pm}$. 
The target variable to learn is the time derivative, at each collocation point $x$, estimated with forward FD as:
\begin{equation}
    \rho_t(x,t)=\frac{\partial \rho(x,t)}{\partial t}= \frac{\rho(x,t+dt)-\rho(x,t)}{dt}.
\end{equation}
The effective relevance weights $W_r(\cdot)$ of the features, as obtained by using ARD, are $W_r(\rho)=0.25, \, W_r(\rho_x)=0.22,\, W_r(\rho_{xx})=0.14,\, W_r(\rho_{xxx})=0.04,\, W_r(I^{\!+})=0.18,\, W_r(I^{\!-})=0.12,\, W_r(x)=0.27$. As can be noted, the third derivative is the least important feature, and we thus decided to disregard it; the \textit{most} important feature is the space $x$ highlighting that the IPDE is not translational invariant. The dependency on $x$ implicitly captures the location of the source term $Q$ representing the resetting of the state of the agents at the origin as in the Fokker-Planck-type equation \eqref{eq:FP}.
 Note that the integral features $I^{\!\pm}$ are also important. This is in line with the theoretical results regarding the Fokker-Planck-type equation \eqref{eq:FP}.

\subsubsection{Learning the mesoscopic IPDE operator via neural networks}
For learning the right-hand-side operator of the IPDE, we have considered as input the relevant features found in the previous section.
\begin{figure}[htb!]
    \centering
    \subfigure[]{\includegraphics[width=0.41 \textwidth]{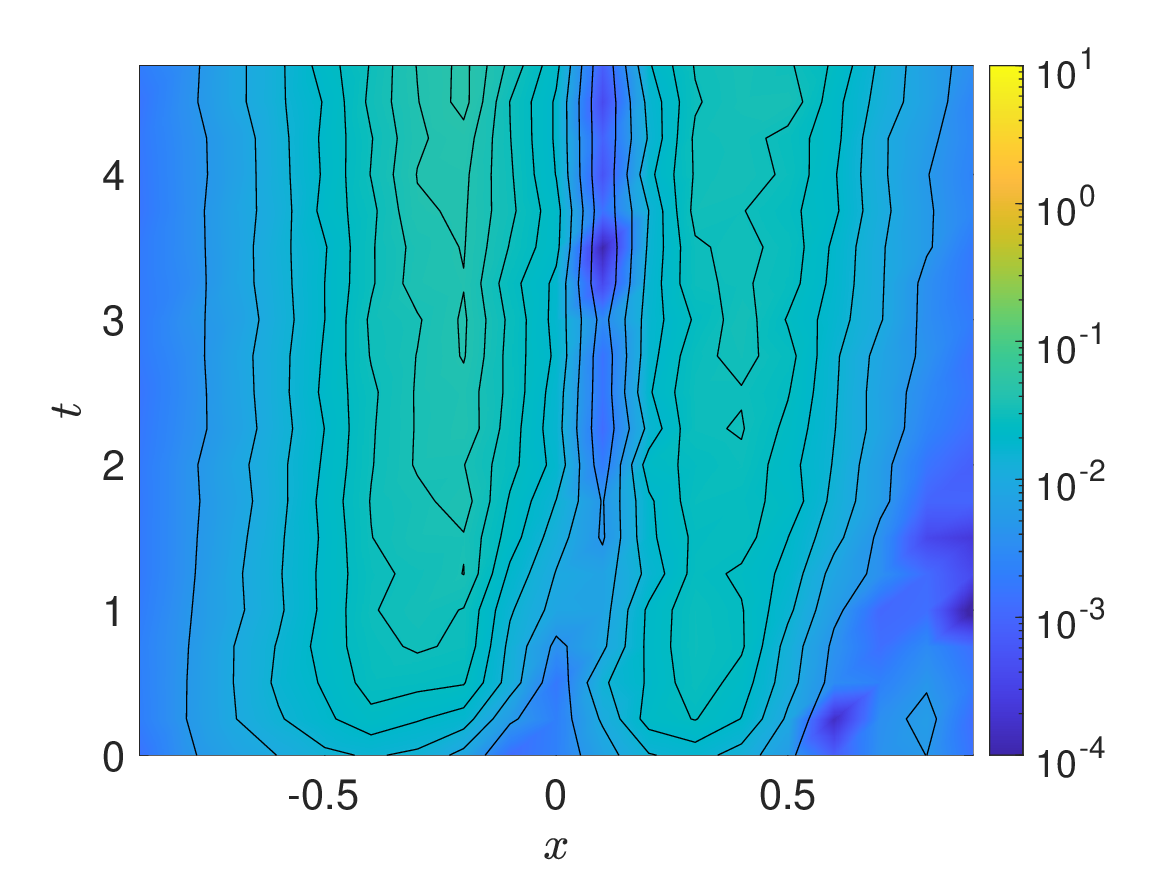}
    \label{fig:time_evo_comparison_a}
    }
    \subfigure[]{\includegraphics[width=0.41 \textwidth]{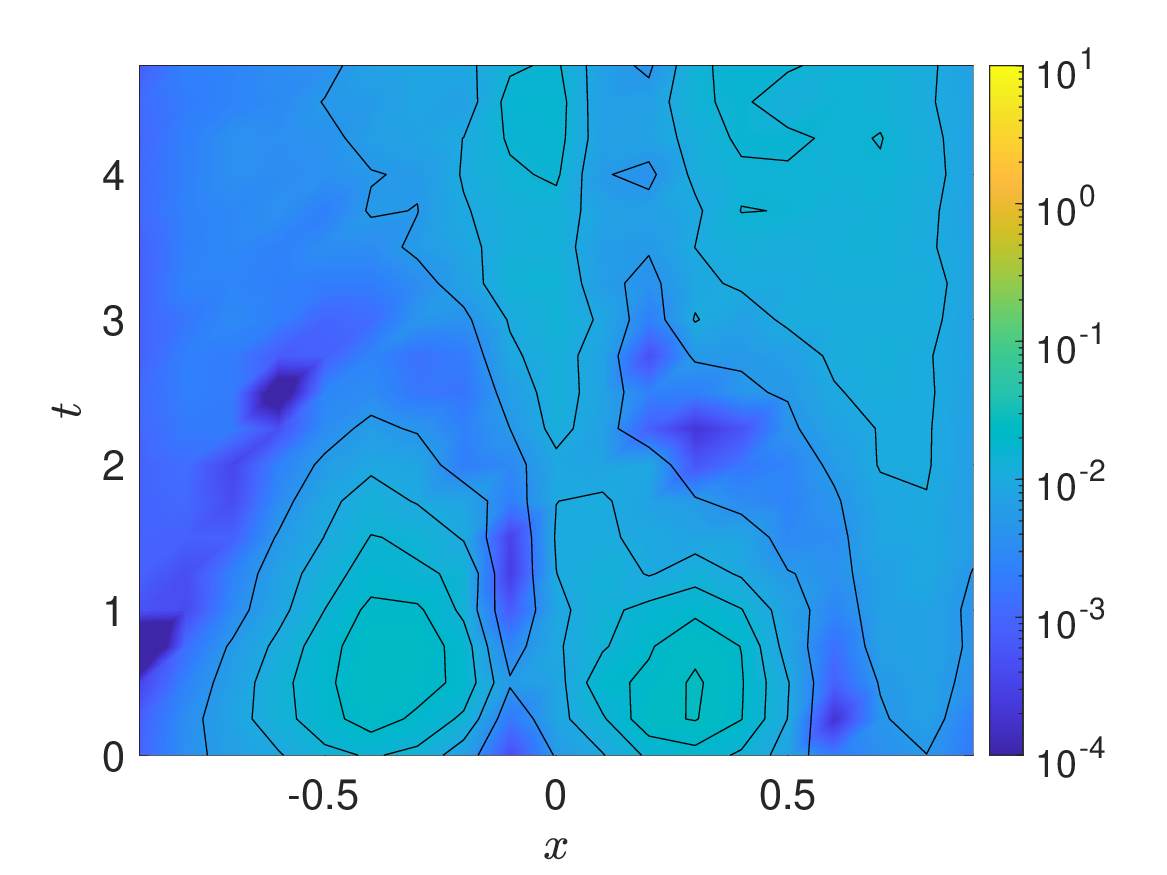}
    \label{fig:time_evo_comparison_b}
    }
    \subfigure[]{\includegraphics[width=0.41 \textwidth]{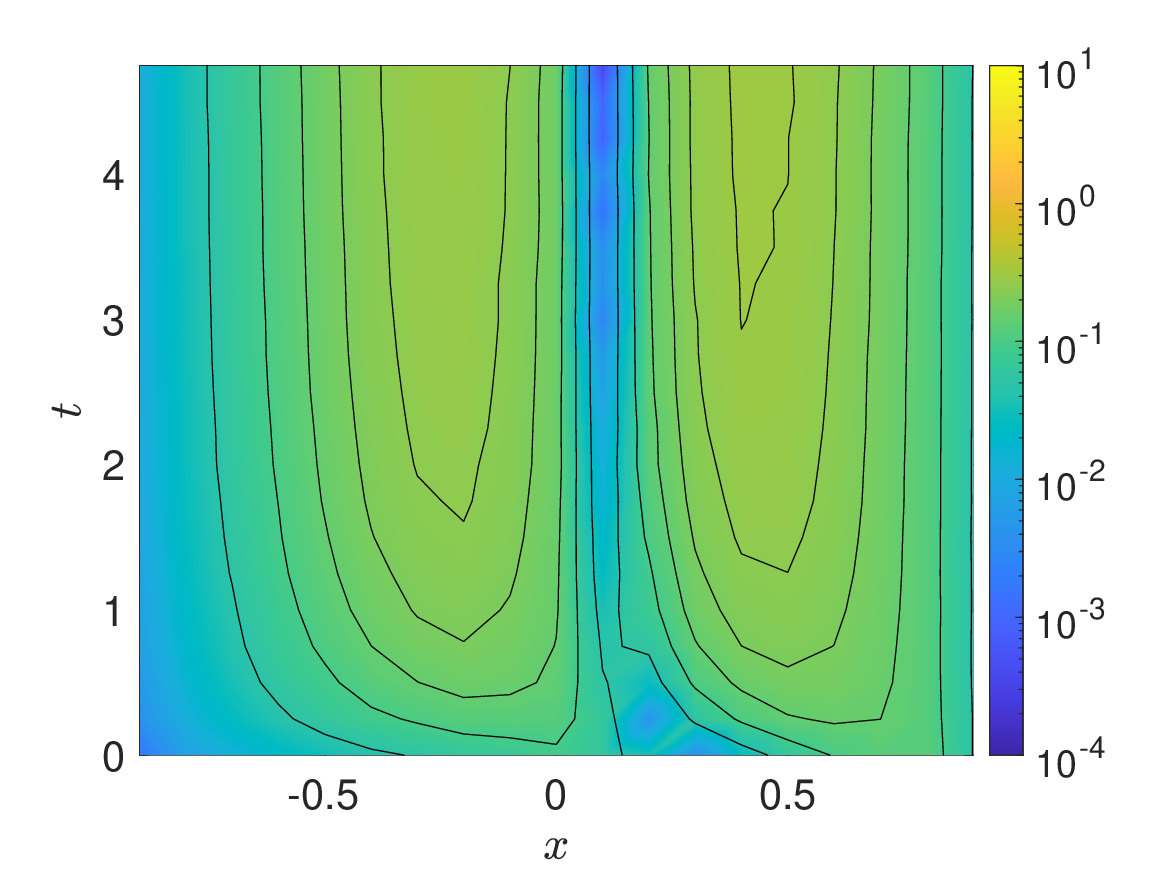}
    \label{fig:time_evo_comparison_c}
    }
    \subfigure[]{
    \includegraphics[width=0.41 \textwidth]{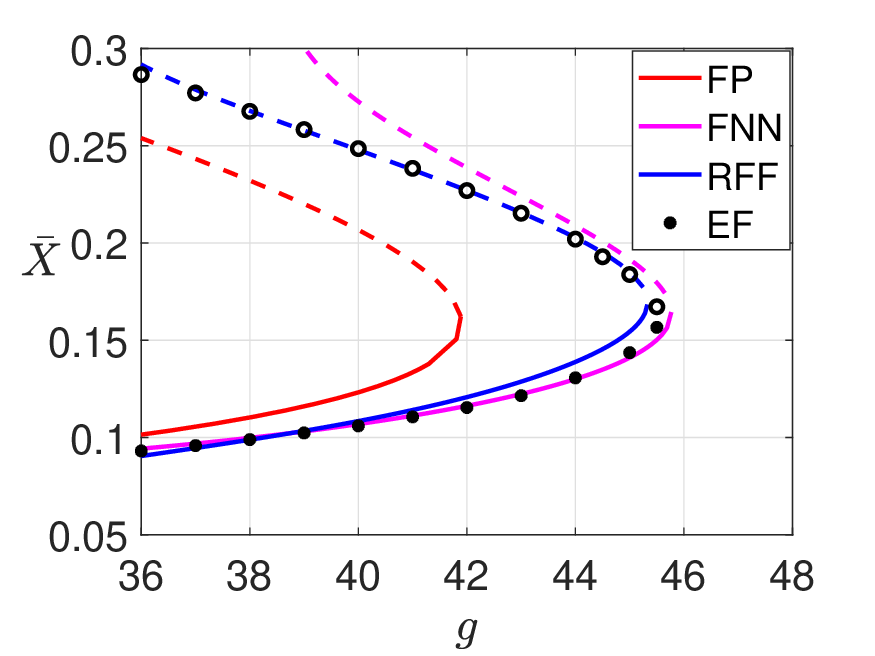}
    \label{fig:BD}
    }
    \caption{(Color                  
              version online) Simulation results for $g=40$ of the various models (analytically derived Fokker-Planck (FP)-type, and the surrogate mesoscopic IPDEs using a two-hidden layer feedforward neural network (FNN) and a single-hidden layer Random Fourier features network (RFF); for comparison purposes, we present here a single arbitrary ABM trajectory, so that all simulations start from the same initial profile, corresponding to $\bar{X}\approx -0.02$, $R^{\!+}\approx0.034$ and $R^{\!-}\approx0.042$. (a)-(c) Absolute error in space and time w.r.t. the ``ground truth'' ABM simulations: (a) model learned with FNN, (b) model learned with RFF, (c) analytically derived FP PDE \eqref{eq:FP},
              (d) Reconstructed bifurcation diagram w.r.t. $g$ obtained with the mesoscopic IPDE surrogate models using a two-hidden layer Feedforward Neural Network (FNN) and a single-hidden layer Random Fourier Features network (RFF); the one computed from the Integro Fokker-Planck (FP), see Eq.\eqref{eq:FP}, and the one constructed with the Equation-free (EF) approach are also given\cite{Siettos2012equation,patsatzis2023data}. Dashed lines (open circles) represent the unstable branches.              \label{fig:time_evo_comparison}}
\end{figure}
Hence, we used the following (black-box) mesoscopic model for the dynamic evolution of the density:
\begin{equation}
    \frac{\partial \rho(x,t)}{\partial t}=F(x,\rho(x,t),I^{+},I^{-},\frac{\partial \rho(x,t)}{\partial x},\frac{\partial^2 \rho(x,t)}{\partial x^2}; g)
    \label{eq:IPDE_ABM}
\end{equation}
The parameter $g$ is also given as input to the FNN. Here, for learning the RHS of the black-box IPDE \eqref{eq:IPDE_ABM}, we implemented two different structures, namely (a) a feedforward neural network (FNN) with two hidden layers, both with 16 Neurons employing a hyperbolic tangent activation function; and (b) a single hidden layer Random Fourier Feature network (RFFN) with $350$ neurons. The two network structures are described in more detail in Appendices \ref{subsec:FNN} and \ref{subsec:RPNN}.
\begin{table}[htb!]
    \centering
     \caption{Learning performance on the training set, in the terms of Mean Absolute Error (MAE), Mean Squared Error (MSE) and the Pearson Correlation/Regression coefficient (R), for the two alternative machine learning approach: Random Fourier Features (RFF) and Feedforward Neural Network (FNN). We highlighted the time-efficient performance of RFF.}
    \begin{tabular}{|c||c|c|c|c|}
        \hline
        network & MAE & MSE & R & training time (s)\\
        \hline\hline
        RFF &  1.09E$-$04 & 2.53E$-$08 & 0.9873 & \cellcolor{LightCyan}26.06\\
        \hline
        FNN & 1.10E$-$04 & 2.60E$-$08 & 0.9866 & 1488.33\\
        \hline
    \end{tabular}
   \label{tab:regressionPDE_RFF_FNN}
\end{table}
For training the FNN, we used the Deep Learning toolbox of \texttt{Matlab} 2021a; for the training of the RFFN, we used the built-in function (\texttt{pinv})  of \texttt{Matlab} implementing the Moore-Penrose pseudo-inverse, with regularization tolerance set to 1E$-$06.
All computations were carried on a 2 GHz Intel Core i5 quad-core with a 16 GB 3733 MHz LPDDR4X memory.
In Table \ref{tab:regressionPDE_RFF_FNN}, we show the performance of the two schemes on the test set, in terms of the Mean Absolute Error (MAE), Mean Squared Error (MSE), and Regression coefficient $R$. The main notable difference between the two schemes is in the computational time needed to perform the training, since remarkably, the training of the RFFN turned out to be at least 50 times faster than the one required for the deep-learning scheme.
Upon training, we integrated in time the black-box mesoscopic IPDE given by (\ref{eq:IPDE_ABM}) using the explicit Heun's Runge-Kutta scheme.%
Note that the analytical Fokker-Planck-type equation of Omurtag and Sirovich, was integrated in time using a conservative Lax-Wendroff scheme. For all simulations, we have used homogeneous Dirichlet boundary conditions. The result of integration in the time window $[0,5]$ is depicted in Figures~\ref{fig:time_evo_comparison_a}-\ref{fig:time_evo_comparison_c}. As it is shown there, the spatiotemporal approximation error, w.r.t the actual mesoscopic distribution resulting from the ABM simulations, of the FNN and RFFN is lower than the FP one.
It is also worth noting, that the Fokker-Planck-type equation was constructed to be conservative, while the two learned surrogates are not forced to satisfy the normalization property. For all practical purposes, however, the integral of the approximated solution remains in the narrow range $[0.998,1.002]$; an explicit built-in constraint or a penalty are worthy of future exploration.

\subsubsection{Bifurcation analysis of the mesoscopic IPDE}

To locate the tipping point, we have performed bifurcation analysis, using both ML-identified IPDE surrogates, as discussed in Section \ref{sec:locate}. Furthermore, we compared the derived bifurcation diagram(s) and tipping point(s) with what was obtained in \cite{Siettos2012equation,patsatzis2023data} using the EF approach (see in the Appendix \ref{sec:coarse_timestepper} for a very brief description of the EF approach). As shown in Fig.~\ref{fig:BD} the two ML schemes approximate visually accurately the location of the tipping point in parameter space. However, the FNN scheme fails to trace accurately the actual coarse-scale unstable branch, near which simulations blow up extremely fast. More precisely, the analytical Fokker-Planck-type equation predicts the tipping point at $g^*=41.90$ with corresponding steady-state $\bar{X}^*=0.1607$ and the EF at $g^{*}= 45.60$ and $\bar{X}^* =0.1627$; our FNN predictions are at $g^{*}= 45.77$ and $\bar{X}^*=0.1644$, the RFF ones at $g^{*}=45.34$ and $\bar{X}^* =0.1684$.

\subsection{Tipping points via the learned mean-field SDE}
\label{sec:learningSDE}
\subsubsection{Data collection and preprocessing}
\label{sec:data_collection}
In this section we describe how we collected data, specifically targeted to the neighborhood of the tipping point, for the purpose of learning a parametric mean-field SDE. 
ABM simulations were performed using $N=50,000$ agents, with $\nu^{\!\pm}=20$, $\epsilon^{\!+}=0.075$, $\epsilon^{\!-}=-0.072$, $\gamma=1$. We selected 11 equally-spaced values of the mimetic strength $g$, which is used as bifurcation parameter, in the range $g \in [42,47]$.
We gathered at each time stamp the state of every agent ($X_i$), as well as the overall buying/selling rates ($R^{+}$ and $R^{-}$). For convenience, we use the vector $\bm{s} = (X_1, X_2, \cdots, X_{n}, R^{+}, R^{-})^T$, and also denote the mean preference value of the agent state as:
\begin{equation*}
    \bar{X} = \frac{1}{n} \sum_{i=1}^{n} X_i.
\end{equation*}

To sample trajectories that populate the state space (in terms of $\bar{X}$, across multiple values of the parameter $g$) the following protocol was used: we start by running one trajectory of the full ABM for \hbox{$g = 47$}, which ultimately leads to an explosion. This trajectory, was initialized by sampling the agents from a triangular distribution $p(X)$, as implemented in python, (with lower limit $-1$, upper limit $1$, and mode $-0.6$), corresponding to a value $\bar{X}\approx -0.2$. This trajectory was stopped the simulations using the termination condition $\bar{X} \leq >0.4$, indicative of incipient explosion. We selected 25 distinct instances of the state of agents along the stochastic trajectory, corresponding to 25 different values of $\bar{X}$ in the range $[-0.02,0.32]$.

Then, for each value of the parameter $g$, we simulated a total of 50 new stochastic trajectories, two for each distinct initial condition.

As previously done in \cite{liu2015equation}, in order to find the data-based 1D coarse observable $\psi_1$, the DMs algorithm is carried out on 39 intermediate coarse variables. Thus, w We set up $37$ percentile points $p_1, p_2, \cdots, p_{37}$ referring to our discretization of the cumulative distribution function (cdf) of the agents' preference state. For each $p_i$, we computed its quantile function value $Q(p_i)$, where the quantile function $Q(\cdot)$ is defined as the inverse function of the cdf $F_X$ of the random variable $X$.
The first 19 percentile values $p_1, p_2, \cdots, p_{18}, p_{19}$ are set as 0.0005, 0.001, 0.002, 0.003, 0.004, 0.005, 0.0075, 0.01, 0.02, 0.03, 0.04, 0.05, 0.075, 0.1, 0.15, 0.2, 0.3, 0.4, 0.5. The last 18 probability values are set to be symmetric with the first 18, i.e. $p_i = 1 - p_{38 - i}, \qquad \forall i =20, 21, \cdots, 37$;
The remaining two coarse variable are the overall buying and selling rates.
Therefore, the full state $\bm{s}$ has a coarse 39-dimensional representation \hbox{$\bm{s'} = (Q(p_1), Q(p_2), \cdots, Q(p_{37}), R^{+}, R^{-})^T$}. 

\subsubsection{Macroscopic physical observables and latent data-driven observables via DMs}
\begin{figure}[ht!]
    \centering
    \subfigure[]{ \includegraphics[width=0.35\textwidth]{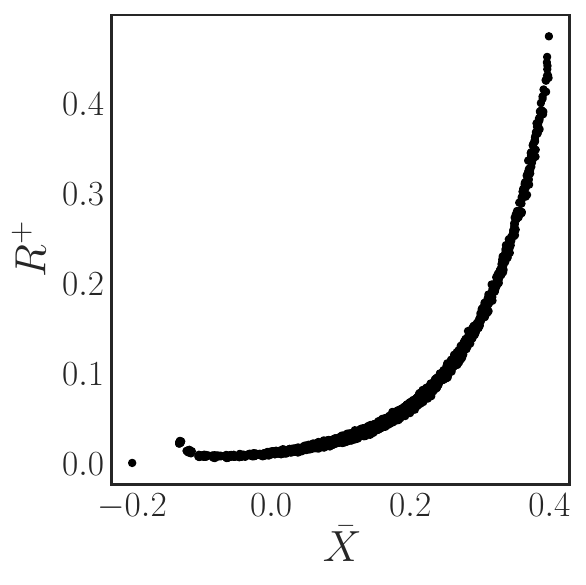}
    \label{fig:dmap_ev_a}}
    \subfigure[] {\includegraphics[width=0.35\textwidth]{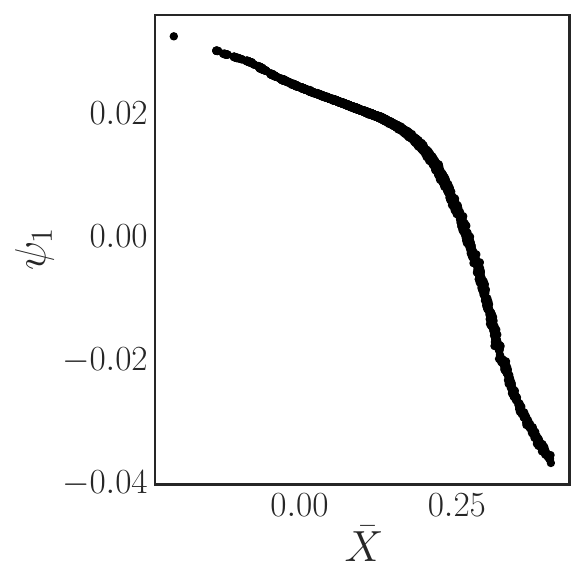}
    \label{fig:dmap_ev_b}} \\
    \subfigure[]{\includegraphics[trim={1cm 0cm 0cm 1.5cm},clip,width=0.4 \textwidth]{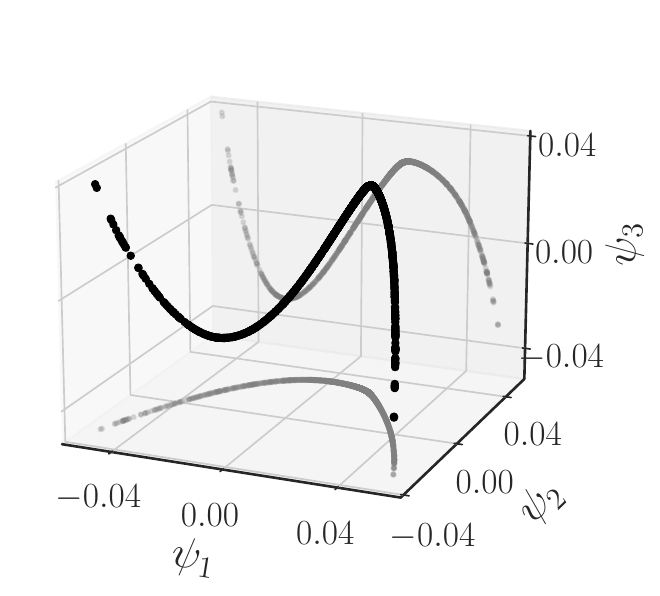}
    \label{fig:dmap_ev_c}}  
    \subfigure[] {\includegraphics[width=0.35 \textwidth]{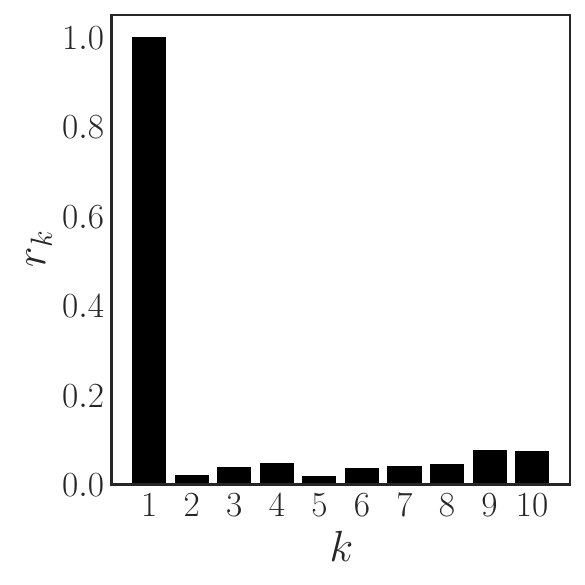}
    \label{fig:dmap_ev_d}}
    \caption{ABM-based-simulation- and Diffusion Maps- (DMs) driven observables. (a) the buying rate $R^{\!+}$ plotted against the mean preference state ($\bar{X}$). (b) The first DMs coordinate $\psi_1$ is plotted against the mean preference state $\bar{X}$. (c) The first three leading DMs coordinates $\psi_1,\psi_2,\psi_3$ plotted against each other, indicating that $\psi_2$ and $\psi_3$ are functions (harmonics) of $\psi_1$; projection to pairs of the eigenvectors are shown with gray color. (d) The estimated residual $r_{k}$ based on the local linear regression algorithm \cite{Dsilva2018parsimonious} indicates that $\psi_1$ is sufficient to parametrize the one-dimensional latent-space, and that the remaining eigenvectors are functions (harmonics) of $\psi_1$. \label{fig:dmap_ev}}
\end{figure}

An immediate physical candidate observable is the first moment $\bar{X}$ of the agent distribution function (as also shown in \cite{liu2015equation}). As simulations of the ABM show (see Figure~\ref{fig:dmap_ev_a}), this mean preference state $\bar{X}$, is one-to-one with another physical meaningful observable, the buying rate $R^{+}$.

We also used the Diffusion Maps (DMs) algorithm, to discover \textit{data-driven} macroscopic observables. In our case, DMs applied to collected data (see Section \ref{sec:data_collection} of the Appendix for a detailed description of how the data were collected), discovers a 1D latent variable $\psi_1$ that is itself one-to-one with $\bar{X}$, see Figure~\ref{fig:dmap_ev_b}. Visual inspection of the first three DMs eigenvectors shown in Figure~\ref{fig:dmap_ev_c} suggests that are \textit{harmonics} (functions) of $\psi_1$. The local-linear regression algorithm proposed in \cite{Dsilva2018parsimonious} was applied to make sure that all the higher eigenvectors can be expressed as local-linear combinations of $\psi_1$ and thus they do not span independent directions. Figure~\ref{fig:dmap_ev_d} illustrates that the normalized leave-one-out error, denoted as $r_k$, is small for $\psi_2,\ldots,\psi_{10}$ suggesting they are all dependent/harmonics of $\psi_1$.

Therefore, any of the three macroscopic observables (two physical and one data-driven) can be interchangeably used to study the collective behavior of the model.

\begin{figure}[ht!]
    \centering
    \subfigure[]{\includegraphics[width=5cm]{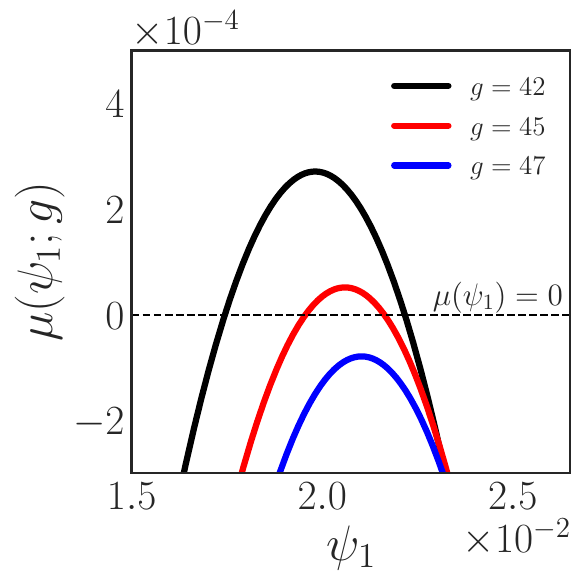}
    \label{fig:sde_psi1_a}}
    \subfigure[]{\includegraphics[width=5cm]{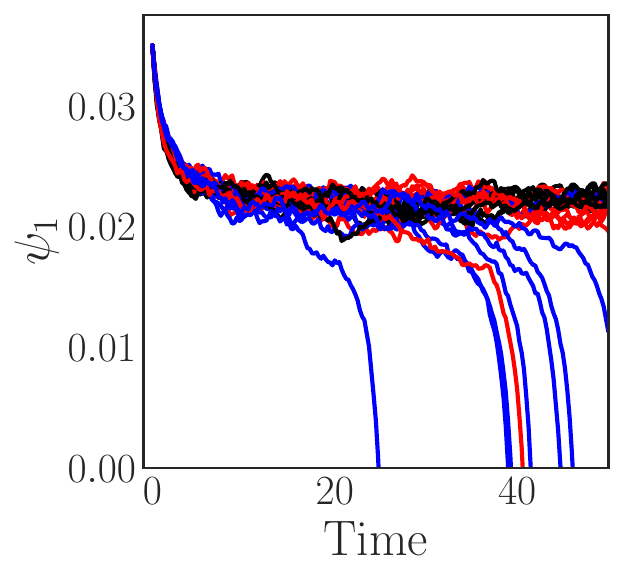}
    \label{fig:sde_psi1_b}}
    \subfigure[]{
\includegraphics[height=4.5cm]{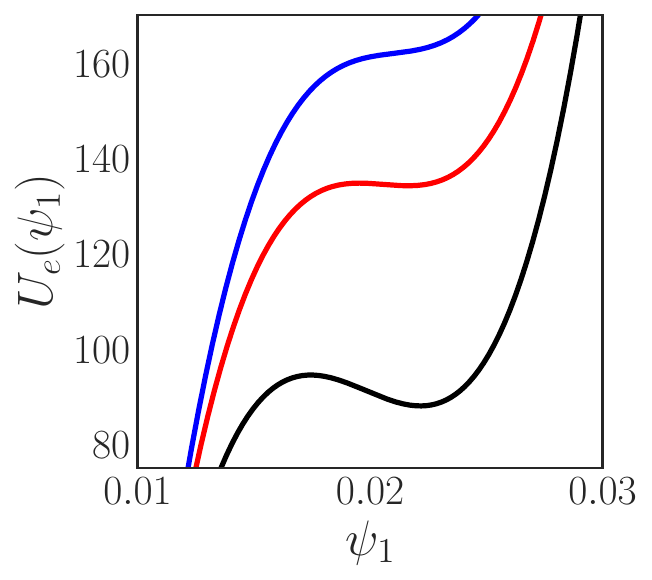}
\label{fig:sde_psi1_c}}
    \subfigure[]{\includegraphics[height=4.55cm]{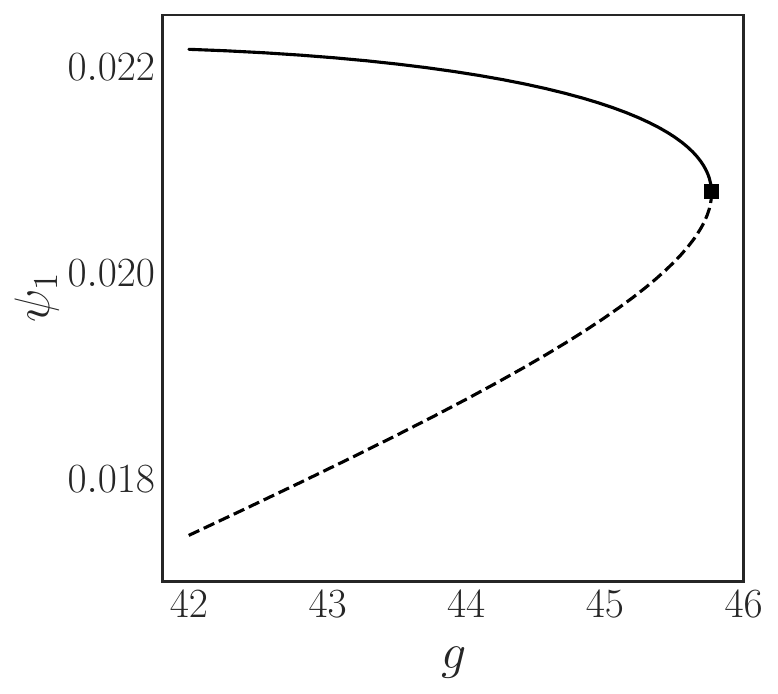}
    \label{fig:sde_psi1_d}}
  \subfigure[]{\includegraphics[height=4.62cm]{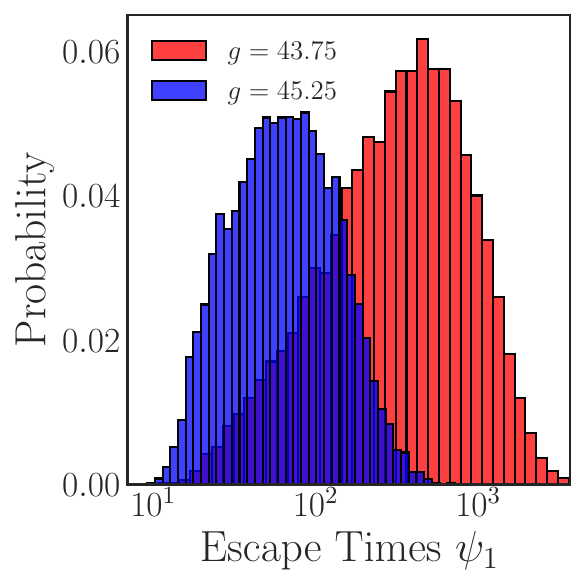}
  \label{fig:sde_psi1_e}}
   \subfigure[]{\includegraphics[height=4.65cm]{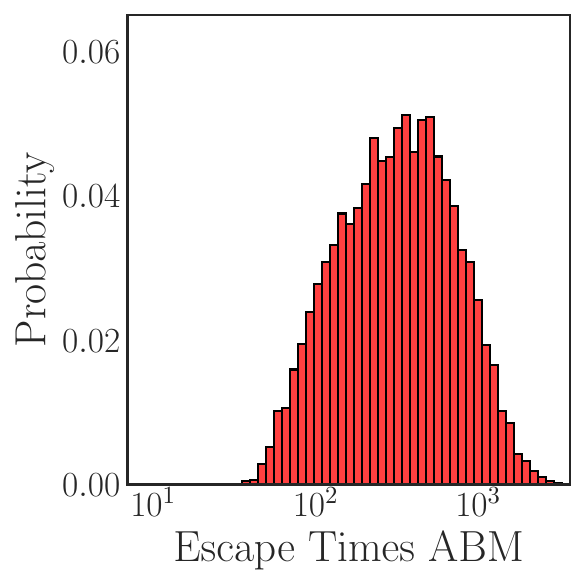}
   \label{fig:sde_psi1_f}}
\caption{(Color version online). The surrogate mean-field macroscopic SDE. (a) The identified drift $\mu(\psi_1;g)$ vs. the DMs coordinate $\psi_1$ for $g=42,45$ and $g=47$. 
(b) Individual paths of the SDE model trained based on $\psi_1$ for the three different values of $g$ (each represented with a different color). (c) The effective potential $U_e(\cdot)$ for the three values of the parameter $g$. (d) The constructed bifurcation diagram based on the drift component of the identified mean-field macrosocpic SDE in $\psi_1$. The bifurcation point is marked with a black square. (e-f) Histograms of escape times obtained with simulations of 10,000 stochastic trajectories for (e) the SDE model trained on $\psi_1$ for $g=45.25$ (blue histogram) and $g=43.75$ (red histogram), (f) the full ABM at $g=45.25$. \label{fig:sde_psi1}}  
\end{figure}

\subsubsection{Learning the mean-field SDE and performing bifurcation analysis}
\label{sec:BF}
Here, for our illustrations, we learned parameter$-$dependent SDEs for three coarse variables, namely the physical variables $\bar{X}$, $R^{\!+}$, and the DMs coordinate $\psi_1$. We first report the results for the identified SDEs in terms of the data-based DMs coordinate $\psi_1$ and then in Section \ref{sec:sde_add_results} we report the results for the identified SDEs with respect to the physically-based $\bar{X}$ and $R^{\!+}$ variables.\par

For all three networks, we have used 5 hidden layers with 32 neurons for each layer. The activation functions for the \textit{drift} neural network were selected as \textit{tanh} and for the diffusivity as \textit{softplus}. 

The data were split 90$|$10 into training$|$validation subset.

In Figure~\ref{fig:sde_psi1_a} the drift coefficient $\mu(\psi_1;g)$ of the SDE is plotted against the identified macroscopic variable $\psi_1$ for three values of the parameter $g$. Figure~\ref{fig:sde_psi1_a} indicates that for values of $g\le 45$ two steady states appear (one stable and one unstable), but for $g>45$ the steady states disappear. The steady states occur when the drift coefficient crosses $\mu(\psi)=0$. This observation indicates that a saddle-node bifurcation occurs. The identified diffusivity coefficient does not appear to depend on the state variable nor on the parameter. 

Given this trained macroscopic SDE surrogate, as shown in Figure \ref{fig:sde_psi1_b}, we computed multiple trajectories from the same initial condition for $g=42,45$ and $47$. The obtained paths also suggest the existence of a stable steady state for $\psi_1 \sim 0.02$ for $g=42$ and $g=45$. For $g=42$ it appears that the trajectories evolve towards a stable steady-state and then spend time around the steady state. For $g=45$ the trajectories are attracted towards the stable steady state but some of those \textit{escape} beyond the unstable steady state and subsequently explode, since the stable and unstable steady states are \textit{closer}. For $g=47$ the trajectories appear to be initially evolving towards the same phase-space location, but then very quickly blow-up to infinity.\par

The effective potential $U_e(\psi_1)$ (see Eq. \eqref{eq:effective_potential} ), was also computed based on the identified drift and diffusivity functions from the SDE. In Figure~\ref{fig:sde_psi1_d}, we illustrate the $U_e(\psi)$ for $g=42,45$ and $g=47$. The stable steady states are minima of the effective potential $U_e(\psi_1)$. The unstable steady states are transition states (saddles, and in 1D local maxima). For $g=47$ the effective potential $U_e(\psi_1)$ is monotonic (there is no local minimum) indicating the absence of a stable steady state.\par  

The drift term (deterministic component) of the identified dynamics was used to construct the bifurcation diagram with AUTO (see Figure \ref{fig:sde_psi1_d}). A saddle-node bifurcation was identified for $g^* = 45.77$ where $\psi_1^{\star} = 0.021$. The estimated critical parameter value from the SDE is in agreement with our previous work ($g \approx 45.60$ \cite{Siettos2012equation}).

\subsubsection{Rare-event analysis of  catastrophic shifts (financial ``bubbles'') via the identified mean-field SDE} \label{sec:MFPT_exp}

Given the identified steady states at a fixed value of $g$ we performed escape time computations. For $g=45.25$, we estimated the average escape time needed for a trajectory initiated at the stable steady state to reach $\bar{X} = 0.3$, i.e. sufficiently above the unstable branch. As shown in Figure \ref{fig:dmap_ev}, $\psi_1$ and $R^{+}$ are effectively one-to-one with $\bar{X}$, and we can easily find the corresponding critical values for $\psi_1=-0.01$ (flipped) and $R^{+}=0.16$. We now report a comparison between the escape times of an SDE identified based on the DMs coordinate $\psi_1$ and those of the full ABM. In Appendix \ref{sec:sde_add_results} we also report the escape times of the SDE for $\bar{X}$ and $R^{\!+}$ observables. To estimate these escape times we sampled a large number ($10,000$ in our case) of trajectories. In Section \ref{sec:quad_appendix} of the Appendix, we also include the escape time computation by using the closed-form formula for the 1D case. The computation there was performed numerically by using quadrature and the milestoning approach \cite{bellorivas2016simulations}.\par 
In Figure~\ref{fig:sde_psi1_e} the histograms of the escape times for the SDE trained on $\psi_1$ and the full ABM model for $g=45.25$ are shown. In Figure \ref{fig:sde_psi1_f}, for the identified SDE in $\psi_1$ we also illustrate the empirical histogram of escape times for $g=43.75$. In Table \ref{tab:escape_times_tab} the estimated values for the mean and standard deviation are reported for these two models.\par 
As shown, the mean escape time of the full ABM is a factor of five larger than that estimated by the simplified SDE model in $\psi_1$ for $g=45.25$ (still within an order of magnitude). The SDE model for $g=43.75$ gives an escape time comparable to the one of the ABM for $g=45.25$. 
Given that the escape times change exponentially with respect to the parameter distance from the actual tipping point, a small error in the identified tipping point easily leads to large (exponential) discrepancies in the estimated escape times.

\begin{table}
\centering
\caption{Means and Standard deviations of escape times as computed with temporal simulations from the SDE trained on the Diffusion Map variable $\psi_1$ for $g=45.25$, $g=43.75$ and the ABM at $g=45.25$ respectively. }
    \begin{tabular}{|c||c|c|c|}
        \hline
        Models & SDE at $g=45.25$ & SDE at $g=43.75$ & ABM \\ \hline\hline
        Mean Escape Time & 84.07 & 480.92 & 434.00
        \\ \hline
             Escape Time Standard deviation & 68.91 & 454.83 &  363.64 
        \\ \hline 
        \end{tabular}
    \label{tab:escape_times_tab}
\end{table}

\paragraph{Computational cost}
We compared the computational cost required to estimate escape times with temporal simulations, through the full ABM and the identified mean-field SDE. To fairly compare the computational costs, we computed the escape times with the ABM for $g=45.25$, and that of the SDE for $g=43.75$, since the two distributions of the escape times are more comparable.

The estimation in both cases was conducted on \textit{Rockfish} (a community-shared cluster at Johns Hopkins University) by using a single core with 4GM RAM. The escape time computation of the ABM was done with \textit{job arrays} to decrease the large computational time (by using 4GM RAM and a single core per job). Using job arrays allowed us to parallelize the computation of the escape times for the ABM and still track the time of computation for each trajectory separately. The computations with the SDE were performed using a single script and conducted in series. In this case, since the computational time was much smaller, using job arrays was not necessary.\par 
For the 10,000 sampled stochastic trajectories, the total computational time needed to perform the escape time computations for the identified coarse SDE in $\psi_1$ was $33.56 \text{min}$, the average time per trajectory, $3.36 \times 10^{-3} \text{min}$ and the approximated mean time per function evaluation $1.74 \times 10^{-6} \text{min}$. The mean time per function evaluation was approximated as the ratio $\dfrac{\text{mean time per trajectory}}{\text{mean number of iterations}}$; the exact computation would have required us to track the total number of steps per trajectory, which we avoided.\par 
For the ABM, the total computational time needed for the escape time computations was $26732.83 \text{min}$ ($18.56$ days), the mean time per trajectory was $2.67 \text{min}$, and the approximated mean time per function evaluation was $1.53 \times 10^{-3} \text{min}$. We reiterate that the total computational time, in this case, was obtained by adding the computational time needed for each trajectory, even though the computation was performed in parallel. 
The total computational time for the escape time computations with the SDE model in $\psi_1$ was $796.56$ times faster than the ABM. A single function evaluation of the identified SDE model was approximately $880$ times faster than the function evaluation of the ABM. This highlights the computational benefits of using the reduced surrogate models in lieu of the full ABM for escape time computations.

\subsection{Additional Results}
\subsubsection{two physical based alternative mean-field SDEs}
\label{sec:sde_add_results}

\begin{figure}[ht!]
    \centering
        \subfigure[]{
\includegraphics[height=5cm]{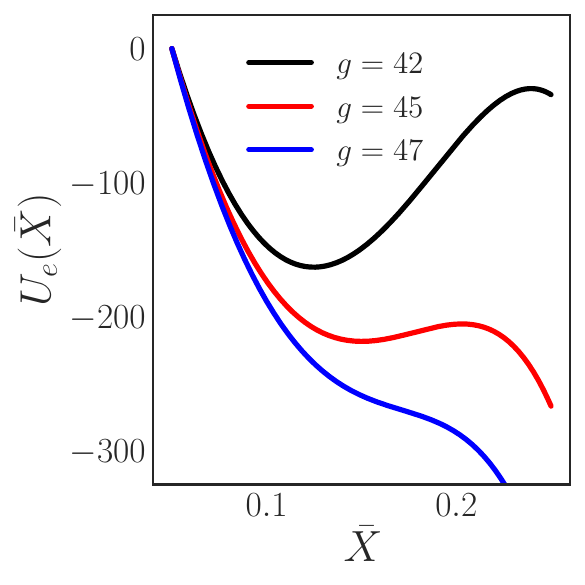}}
    \subfigure[]{
\includegraphics[height=5cm] {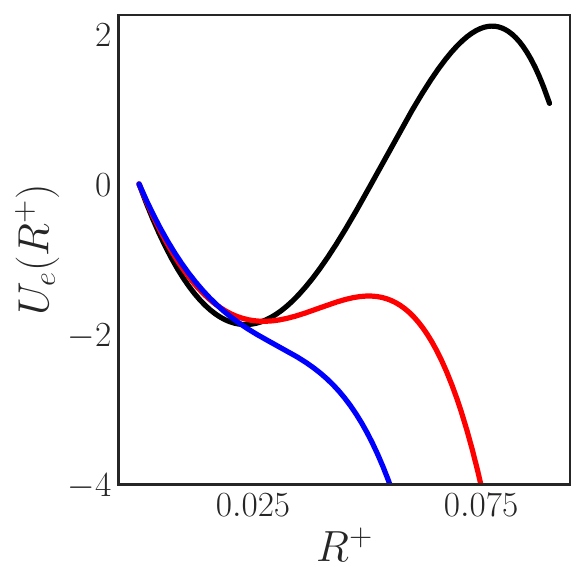}}\\
    \subfigure[]{\includegraphics[height=4.75cm]{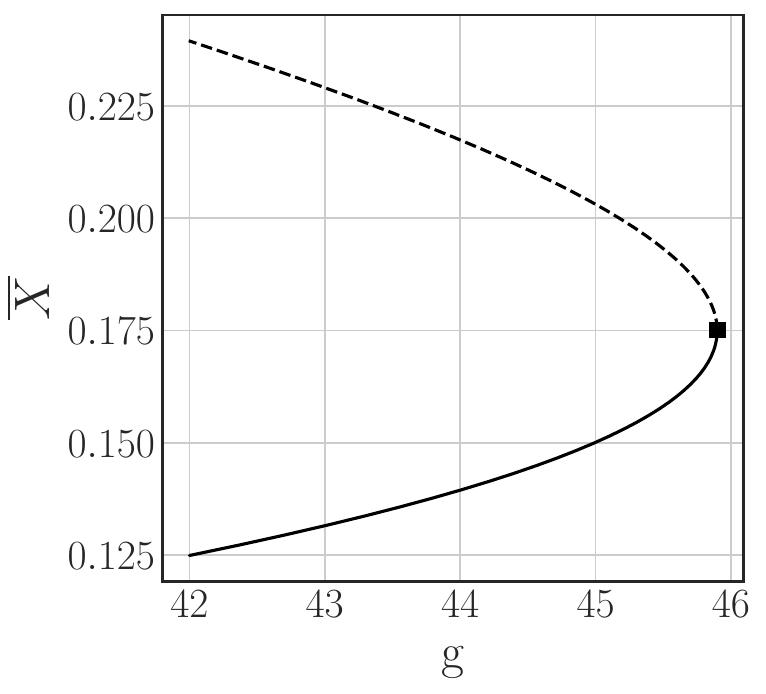}}
    \subfigure[]{\includegraphics[height=4.75cm]{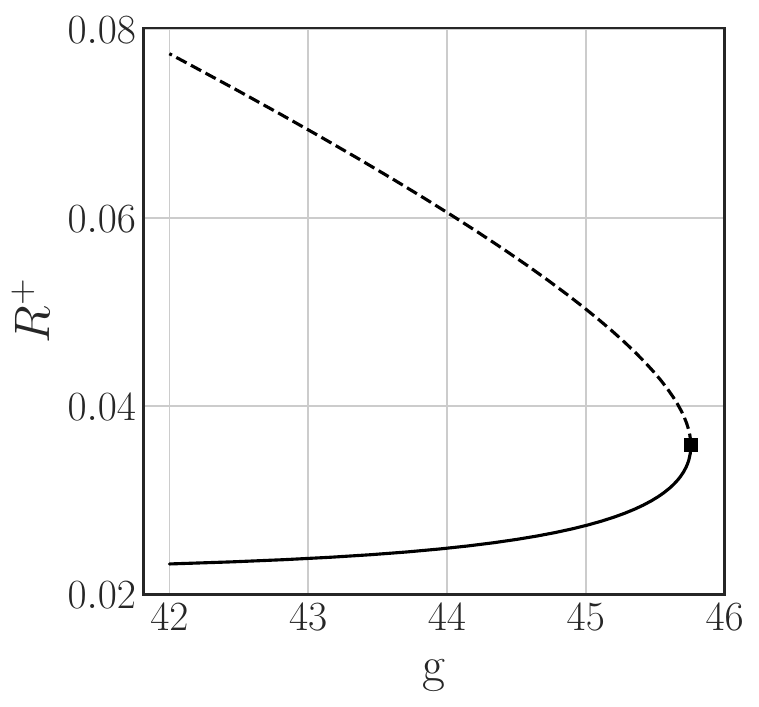}}
    \caption{The effective potential $U_e(\cdot)$ for the two alternative SDE models, in terms of (a) $\bar{X}$, and of (b) $R^{\!+}$ for $g=42,45$ and $g=47$. The constructed bifurcation diagrams are based on the drift component of the identified effective SDEs in terms of (c) $\bar{X}$, and of (d) $R^{+}$. The bifurcation point is marked with a black square.}
    \label{fig:bifurcation_diagram2}
\end{figure}

In this section, we illustrate results obtained for two alternative 1D mean-field SDE surrogates, using as their effective state variable $\bar{X}$ and $R^{+}$, respectively; these were omitted in the main text for brevity.
\begin{table}[ht!]
    \centering
    \caption{Tipping points and stability for different alternative SDE surrogates.}
    \begin{tabular}{|c||c|c|}
        \hline
        Variables & $\bar{X}$ & $R^{+}$ \\ \hline\hline
        Bifurcation parameter $g^*$ & 45.8979 & 45.7570 
        \\ \hline
        Tipping point value $x^*$ & $\bar{X}^* = 0.1751$ & $R^{+*} = 0.003586$ 
        \\ \hline
        Stable region & $\bar{X} < \bar{X}^*$ & $R^{+} < R^{+*}$
        \\ \hline
        Unstable region & $\bar{X} > \bar{X}^*$ & $R^{+} > R^{+*}$
        \\ \hline
    \end{tabular}
    \label{tab:bifurcation_diagram_tab2}
\end{table}
In Figure~\ref{fig:bifurcation_diagram2}(a)-(b) we illustrate the estimated effective potential $U_e(\cdot)$ for each identified 1D SDE surrogate (in terms of $\bar{X}$ and in terms of $R^{+}$) across three values of the mimetic parameter $g=42,45$ and $g=47$.

The constructed bifurcation diagrams from the identified deterministic drift of the two SDE surrogates trained on $\bar{X}$ and $R^{+}$ are shown in Figure~\ref{fig:bifurcation_diagram2}(c)-(d).

The bifurcation point, in both cases, occurs for $g^{*} \approx 45.5$ as shown in Table \ref{tab:bifurcation_diagram_tab2}. The identified bifurcation point (our tipping point) for all three identified SDE models is consistent with the value reported in Liu P. et al \cite{Siettos2012equation,liu2015equation,patsatzis2023data}.

\begin{figure}[ht!]
    \centering
    \subfigure[]{\includegraphics[height=5.1cm]{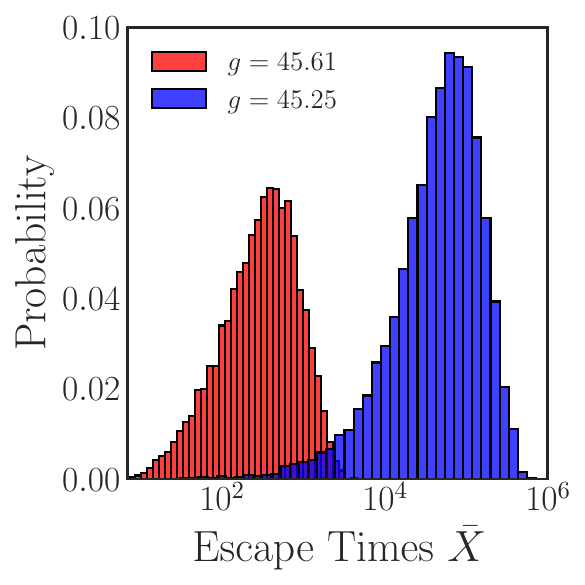}}
    \subfigure[]{\includegraphics[height=5cm]{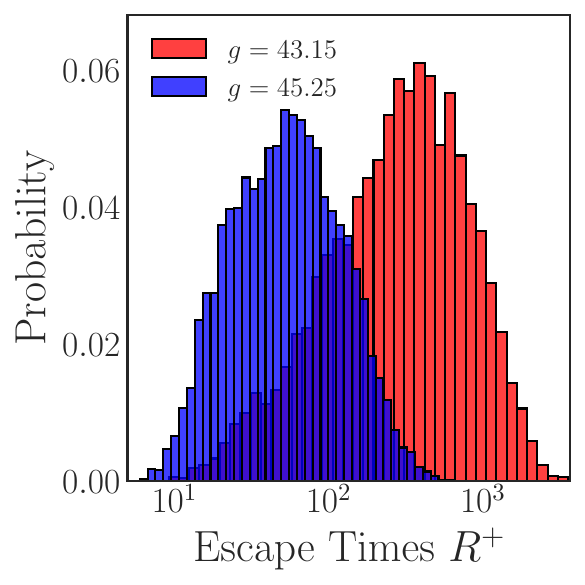}}
    \caption{Histograms of escape times obtained with simulations of 10,000 stochastic trajectories for the SDE models trained on (a) $\bar{X}$, (b) $R^{+}$. \label{fig:escape_times2}}
    \end{figure}

For each of the identified SDEs, in terms of $\bar{X}$ and $R^{+}$ respectively, we estimated escape time distribution by computing $10,000$ stoschastic simulations. Histograms of the obtained escape times for $\bar{X}$ and $R^{\!+}$ are shown in Figure~\ref{fig:escape_times2}(a)-(b), respectively. The mean and standard deviation of the results of the simulations are reported in Table \ref{tab:escape_times_tab2}. Similarly to the computations reported in the main text, for the SDE models in $\bar{X}$ and $R^{+}$ we estimated the mean escape times for $g=45.25$; We also found parameter values of $g$ that provide comparable mean escape time with that of the full ABM for $g=45.25$.

The estimated escape times of the identified SDE in $\bar{X}$ for $g=45.25$
is much larger than any other model (including the ABM). This might suggest that this surrogate model might be unreliable. The surrogate SDE model constructed in $R^{+}$ has an escape time similar to the model trained on $\psi_1$.

    \begin{table}[ht!]
        \centering
        \begin{tabular}{|c||c|c|c|c|}
        \hline
        Models &  $\bar{X}$ for $g=45.25$ & $\bar{X}$ for $g=45.61$ & $R^{+}$ for $g=43.15$ & $R^{+}$ for $g=45.25$  \\ \hline\hline
        Mean & 75627.73 & 468.06 & 436.59 & 73.61  
        \\ \hline 
             Standard deviation & 75198.20 & 456.71 & 420.82 & 63.95  
        \\ \hline
    \end{tabular}
    \caption{Means of escape times distribution and their standard deviations for the identified SDE in terms of $\bar{X}$ and $R^{+}$. \label{tab:escape_times_tab2}}
    \end{table}


\subsubsection{Escape Times by quadrature}
\label{sec:quad_appendix}

In this Section we report results of the alternative approach discussed in Section \ref{sec:quad} to compute escape times for 1D systems. This computation of mean escape times involves quadrature integration for the three different SDE surrogates in $\bar{X},R^{+}, \psi_1$
coordinates, respectively. The transformation is discussed in Section \ref{sec:quad} of the Appendix. The estimated mean escape times for $g=45.25$ are reported in Table \ref{tab:escape_times_quadrature} for the three models. 

\begin{figure}
    \begin{floatrow}
        \capbtabbox[]{
        \centering
        \begin{tabular}{|c||c|c|c|}
        \hline
        Variables & $\bar{X}$ & $R^{+}$ & $\psi_1$ \\ \hline\hline
        Escape times & do not converge & 39.64 & 22.23\\
        \hline
    \end{tabular}
    \vspace{0.3cm}
        }
        {\caption{Escape times computed with the quadrature method for the three SDE models $\bar{X},R^{\!+},\psi_1$}\label{tab:escape_times_quadrature}}
        \capbtabbox[]{
        \centering
        \begin{tabular}{|c||c|c|c|c|}
        \hline
         Model & z  & $\beta$ & $a$ & $b$\\ \hline \hline
         $\sigma^{-1}\bar{X}$ & 43.89 & 2 & 10.00 & 62.71  \\ \hline 
   $\sigma^{-1}R^{\!+}$ & 4.34 & 2 & 0.00 &8.65  \\ \hline 
      $\sigma^{-1}\psi_1$ & 41.87 & 2 & 36.91    & 48.00  \\ \hline
    \end{tabular}
        }
        {\caption{The parameter values used for the escape time computations with the quadrature method across the three transformed  models.\label{tab:table_parameters_quadrature}}
        }
    \end{floatrow}
\end{figure}

As it can be seen from Table \ref{tab:escape_times_quadrature}, the estimated escape times for the SDEs on the $R^{+}$ and $\psi_1$ are comparable to the ones obtained with the Monte-Carlo simulations (reported in Section \ref{sec:MFPT_exp}).  However, the escape times obtained with the SDE trained on $\bar{X}$ did not converge because of overflow errors. The latter is due to numerical issues that arise in the quadrature involving the computation of the exponential of the potential in Eq. \eqref{eq:Juan_G}.

We now discuss in more detail how the parameters $\alpha,\beta$ and $z$, specifying the limits of the integrals \eqref{eq:Juan_G}, were selected for the quadrature integration:
\begin{enumerate}
    \item The parameter $z$ indicates the point where the process is initiated. In our case this was selected to be the stable steady state identified by the drift network for a fixed value of the parameter $g=45.25$. We set the value of the potential for $U_e(\cdot)$ to be equal to zero.
    \item The parameter $\beta$ is estimated based on the diffusivity $\sigma$ of the SDE. Assuming that $\sigma = \sqrt{2\beta^{-1}}$ and thus $\beta = \frac{2}{\sigma^2}.$ Because $\beta$ is being raised to an exponent for the quadrature computations, when $\sigma$ is not close to one, numerical overflow can arise. This is the reason we decided to perform It\^o's transformation to the SDE.
    
    \item $a$ and $b$ are selected based on the shape of the potential. In the cases that the unstable steady state was at larger values (e.g for the SDE trained on $R^{\!+}$ and $\bar{X}$) then the stable steady state $b$ was selected as the coordinate after the unstable steady state that has potential $U_e(\cdot) =0 $. 
    
  In this case, $a$ is set as the point beyond which a trajectory (in a more general sense a particle) is considered of having negligible probability of finding itself. This is due to the fact that the potential energy becomes too large. For the $\psi_1$ the selection of $a$ and $b$ is done in opposite order. To select a value for $a$ or $b$ when they represent this boundary we gradually increased or decreased their values until the computed escape time remains unchanged (converges). 
\end{enumerate}

In Table \ref{tab:table_parameters_quadrature} we report the parameters $z, \beta, a$ and $b$ for the three SDE models. Please note that for all those cases the parameter $\beta$ was set to $\beta =2$ because of the applied It\^o's transformation.

\section{Conclusions}
\label{sec:conclusion}
Predicting the probability of catastrophic shifts, designing control policies for them \cite{Siettos2012equation,patsatzis2023data} and thus eventually preventing them, is one of the biggest challenges of our times. Climate change and extinction of ecosystems, outbreak of pandemics, economical crises, can be attributed to both systematic changes and stochastic perturbations that, close to tipping points, can drive the system abruptly towards another regime that might be catastrophic. Such tipping points are most often associated with underlying bifurcations. Hence, the systematic identification of the mechanisms - types of bifurcations- that govern such shifts, and the quantification of their occurrence probability, is of utmost importance. 
Towards this aim, as real experiments in the large scale can be difficult, or impossible to perform (not least due to ethical reasons) mathematical models and especially high-fidelity large-scale agent-based models are a powerful tool in our arsenal to build informative ``digital twins'' (see also the discussion in \cite{scheffer2003catastrophic,scheffer2010foreseeing}).
However, due to the ``curse of dimensionality'' that pertains to the dynamics of such large-scale ``digital twins'', the above task remains  computationally demanding and challenging.\par 
Here, we proposed a machine-learning based framework to infer tipping points in the emergent dynamics of large-scale agent-based simulators. In particular, we proposed and critically discussed the construction of mesoscopic and coarser/mean field ML surrogates from high-fidelity spatio-temporal data for: (a) the location of bifurcation points and their type, and (b) the quantification of the distribution of the occurrence of the catastrophic shifts.
As our illustration, we used an event-driven stochastic agent-based model  describing the mimetic behavior of traders in a simple financial market. The emergent dynamics of the ABM predict both asset bubbles and crashes. 
For the particular ABM, starting from the continuity equation for the agent probability density function, one can construct analytically a closed form mesoscopic Fokker-Planck-type evolution equation, which can qualitatively (but not quantitatively) predict the tipping point.
While ML surrogates in the form of IPDEs provide some physical insight for the emergent dynamics, they may introduce biases in the accurate numerical bifurcation analysis. Furthermore, they cannot be easily used for the task of uncertainty quantification. On the other hand, surrogate mean-field SDEs offer an attractive and computationally ``cheap`` alternative, that can provide an accurate approximation of the tipping point and can, in addition, naturally inform uncertainty quantification.\par
Clearly, different modelling tasks are best served by different coarse-scale surrogates. This underpins the importance of selecting the right ML modelling approach for different system-level tasks, being conscious of the pros and cons of the different scale surrogates. Therefore further extensions of our framework may include the learning of a more general class of SDEs (e.g. based on L\`evy process); and possibly moving towards learning effective SPDEs or even fractional evolution operators that could lead to more informative surrogate models.\par
Finally, we note, that the proposed methodology may be extended to provide precursors of early warning signals from real-world time-series data. 
Having discovered the type of the bifurcation in the emerging dynamics of the right ``digital twin'', one can track characteristic patterns of fluctuations to construct early warnings of rare events \cite{scheffer2009early}. For example, near a saddle-node bifurcation like the one discovered here, as the slowest coarse eigenvalue approaches zero, the impacts of physical perturbations along the corresponding coarse eigenvector will not decay fast enough, thus leading to an increased variance in the pattern of fluctuations observed.  Their magnitude can provide an estimate of the distance to and/or the probable time-interval before the catastrophe strikes. Other types of hard bifurcations, e.g. subcritical Andronov-Hopf, can give rise to different types of catastrophic shifts and different relevant precursors.

\section*{Declarations}
\subsection*{Authorship contribution statement}
\textbf{G.F.}: Data curation, Formal analysis, Investigation, Methodology, Software, Validation, Visualization, Writing – original draft, Writing – review \& editing.\\
\textbf{N.E.} Data curation, Formal analysis, Investigation, Methodology, Software, Validation, Visualization, Writing – original draft, Writing – review \& editing.\\
\textbf{T.C.} Data curation, Methodology, Software, Writing – original draft.\\
\textbf{J.M.B.R.}:Data curation, Formal Analysis, Investigation, Methodology, Software, Writing – review \& editing.\\
\textbf{C.M.L.}: Data curation, Investigation, Methodology, Software.\\
\textbf{C.S.}: Conceptualization, Formal analysis, Supervision, Investigation, Methodology, Software, Validation, Writing – review \& editing.\\
\textbf{I.G.K.}: Conceptualization, Formal analysis, Supervision, Investigation, Methodology, Validation, Writing – review \& editing.

\subsection*{Data Availability}
Data and/or Software will be publicly available upon publication.

\subsection*{Conflict of Interest}
The authors declare that there is no known competing financial interests or personal relationships that could have appeared to influence the work reported in this paper.

\section*{Acknowledgments}
I.G.K. acknowledges partial support from the US AFOSR FA9550-21-0317 and the US Department of Energy SA22-0052-S001.
C.S. acknowledges partial support from the PNRR MUR, projects PE0000013-Future Artificial Intelligence Research-FAIR \& CN0000013 CN HPC - National Centre for HPC, Big Data and Quantum Computing. C.M.L. received the support of a ``la Caixa'' Foundation Fellowship (ID
100010434), code LCF/BQ/AA19/11720048.

\bibliographystyle{unsrt}
\bibliography{bi.bib}

\appendix
\section{Machine Learning algorithms used}
\label{sec:Preliminaries}

\subsection{Diffusion Maps: a Dimension Reduction Approach}
\label{subsec:DMs}

Diffusion Maps (DMs), an algorithm proposed by Coifman and Lafon in 2006  \cite{Coifman2006diffusion}, is a manifold learning technique capable of discovering linear and non-linear patterns in high-dimensional data: an intrinsic embedding of the low-dimensional manifold on which the data (are assumed to) lie. DMs has been applied on data across many fields including images \cite{talmon2013diffusion}, biological data \cite{angerer2016destiny}, molecular simulation data \cite{chiavazzo2017intrinsic}, etc. 

The algorithm, using a random walk on the available data points (each considered as the node of a graph), discovers the underlying structure of the data geometry by approximating the Laplace-Beltrami operator on the data manifold. In this graph the edges between nodes (data points) represent the probability of transitioning from one data point to another. Starting with the data matrix $\mathbf{X} \in \mathbb{R}^{m \times d}$, where $m$ is the number of data points, and $d$ is the dimension of each data point $x_i$, the DMs algorithm constructs an affinity matrix (kernel matrix) $W$: the entries of $W$ are computed as:

\begin{equation}
    w_{ij}=\text{exp}\biggl(-\frac{||\boldsymbol{x}_i-\boldsymbol{z}_j||^2}{2\epsilon}\biggr).
\end{equation}

where $\epsilon$ is a scale parameter. The metric $|| \cdot ||$ we consider here is the $l_2$ norm.

To make the kernel matrix invariant to the sampling density, and to ensure numerical approximation, of the Laplace Beltrami operator the normalization
\begin{equation}
    \tilde{W}=P^{-1}WP^{-1}, P_{ii}= \sum_{j=1}^m W_{ij}
\end{equation}
is applied. 

Then the matrix $D \in \mathbb{R}^{m \times m}$  is constructed as $D_{ii} = \sum_{j=1}^m \tilde{W}_{ij}$ and the second normalization
\begin{equation}
    A = D^{-1}\tilde{W}
\end{equation}
is applied to obtain the row-stochastic matrix $A$.

We then compute the eigendecomposition of $A$,

\begin{equation}
    A\psi_i = \lambda_i \psi_i,
\end{equation}

with the eigenvectors $\psi_i$ sorted based on the eigenvalues $\lambda_i$. Proper selection of the eigenvectors that span independent eigendirections  (termed non-harmonics) is necessary. The selection of the non-harmonic eigenvectors is here obtained by applying the local linear regression algorithm proposed by Dsilva et al. \cite{Dsilva2018parsimonious} on the computed eigenvectors. This linear regression algorithm by fitting eigenvectors $\psi_i$ (where $i>1$) as local linear functions of previous eigenvectors can detect the non-harmonic eigenvectors. A normalized leave-one-out error, denoted as $r_k$, is used for this selection and quantifies gradually which eigenvectors are independent (non-harmonic) and which are not (harmonic). If the number of non-harmonic eigenvectors is smaller than $d$ the dimensionality reduction is achieved.

\subsection{Feedforward Neural Networks}
\label{subsec:FNN}
Feedforward Neural Networks (FNN) are a class of powerful machine learning tools, characterized by a layered structure of interconnected computing units (neurons), that are nowadays widely used for supervised learning tasks, such as regression, classification, forecasting and model identification.
The great popularity of FNN is due to their capability to approximate any (piece-wise) continuous (multivariate) function, to any desired accuracy, as it is stated in the celebrated universal approximation theorem \cite{Cybenko1989approximation,Hornik1989multilayer,
leshno1993multilayer}.
This implies that any failure of a network must arise from an inadequate choice/calibration of weights and biases or an insufficient number of hidden nodes.\par
Let us consider a FNN with a $N_0$-dimensional input $\bm{y}^{0}$, with $L$ hidden-layers composed by $N_l$ neurons. The output $y^{(l)}_j$ of the $j$-th neuron ($j=1,\dots,N_l$) in the $l$-th layer ($l=1,\dots,L$) consists of an evaluation of the so-called activation function $\psi:\mathbb{R}\mapsto \mathbb{R}$ of a linear combination of neurons' outputs of the previous layer:
\begin{equation}
y^{(l)}_j=\sum_{i=1}^{N_l} \psi(w^{(l)}_{ji}y^{(l-1)}_i+b^{(l)}_i) 
\end{equation}
where the weights $w^{(l)}_{ji}$ are the weights of the connection between neurons $i$ and $j$ belonging to two consecutive layers, and $b^{(l)}_i$s are the so-called biases.

if we denote by $\Phi^{l}:\bm{y}^{(l-1)} \in \mathbb{R}^{N_{l-1}} \mapsto \bm{y}^{l} \in \mathbb{R}^{N_l}$ the map between the $(l-1)$-st layer to the $l$-th layer,  then we can express the output $\bm{y}^{L+1}$ of the network as the composition of all the layer maps:
\begin{equation}
\bm{y}^{(L+1)} = \Phi^{(L+1)} \circ \dots \circ \Phi^{1}(\bm{y}^{(0)}).
\end{equation}

\subsubsection{Training of the FNN}
For supervised learning tasks, the goal is to find an \emph{optimal} (most of the times is just sub-optimal) configuration of weights and biases that minimize a loss/cost function, usually defined as the Mean-Squared error between the current output of the network and the desired output $\bm{d}=(d_1,\dots,d_k,\dots,d_M)$ along a collection of $M$ data samples $\bm{y}^{(0)}_1,\dots,\bm{y}^{(0)}_k,\dots,\bm{y}^{(0)}_M$ that constitute the so-called training set:
\begin{equation}
E=\sum_{k=1}^M ||\bm{d}_k-\bm{y}^{(L+1)}_k||_2^2.
\end{equation}

For our computations we used the Bayesian regularized back-propagation updating the weight values using the Levenberg-Marquardt algorithm (\cite{Hagan1994training}) as implemented in \texttt{Matlab}. Levenberg-Marquardt emulates a second-order Newton-like scheme by computing an approximation of the Hessian matrix $H_{\bm{p}}$ of the loss function as:
\begin{equation}
    H_{\bm{p}}\approx J_{\bm{p}}^T J_{\bm{p}},
\end{equation}
where $J_{\bm{p}}$ is the Jacobian matrix of the network errors with respect to weights and biases. Thus, the vector $\bm{p}$ collecting all trainable parameters is iteratively adjusted as follow:
\begin{equation}
    \bm{p}\leftarrow\bm{p}-(J_{\bm{p}}^TJ_{\bm{p}}+\mu I)^{-1}J_{\bm{p}}^T\bm{e},
\end{equation}
where $\bm{e}$ is a vector of network errors.

\subsection{Random Projection Neural Networks}
\label{subsec:RPNN}
Random Projection Neural Networks (RPNNs) are a family of Artificial neural networks including Random Weigths Neural Network \cite{schmidt1992feed}, Random Vector Functional Links (RVFLs) \cite{pao1992functional,pao1994learning,barron1993universal,igelnik1995stochastic}, Random Fourier Features (RFF) \cite{rahimi2007random,rahimi2008weighted} 
Reservoir Computing/ Echo state networks \cite{verstraeten2007experimental,jaeger2001echo,maass2002real} and Extreme Learning Machines (ELMs) \cite{huang2006extreme}. 
The basic idea behind all these approaches, whose seed can be found in the pioneering work of Rosenblatt back in '60s \cite{rosenblatt1962perceptions}, behind all these approaches is to use FNNs with random fixed weights between the input and the hidden layer(s), random fixed biases for the nodes of the hidden layer(s), and a single trainable linear output layer. Based on that configuration, the output is projected linearly onto the functional subspace spanned by the nonlinear basis functions produced by the hidden layers, and the only remaining unknowns are the weights between the last hidden and the output layer. They are estimated by solving a regularized least squares problem \cite{fabiani2022parsimonious,fabiani2021numerical}. 
The universal approximation property of the RPNNs has been proved in a series of papers (see e.g. \cite{barron1993universal,igelnik1995stochastic,huang2006extreme}).\par
Here for simplicity we consider a feedforward RPNN, with one single hidden layer, one output layer with linear activation function and zero output bias $b_i^{(2)}=0$, that can be written as:
\begin{equation}
    y^{(2)}=\sum_{i=1}^N w_i^{(2)} \psi(\bm{w}_i^{(1)}\bm{y}^{(0)}+b_i^{(1)})
    \label{eq:RPNN}
\end{equation}
and we consider a training set of $M$ points $\bm{y}^{(0)}_1,\dots,\bm{y}^{(0)}_k,\dots,\bm{y}^{(0)}_M$ with desired outputs $\bm{d}=(d_1,\dots,d_k,\dots,d_M)$.
Randomly fixing $\bm{w}_i^{1}$ and $b_i^{(1)}$, we can rewrite \eqref{eq:RPNN} as a linear system in the unknowns $\bm{w}^{(2)}$:
\begin{equation}
    \bm{d}=\bm{w}^{(2)}\Phi
\end{equation}
where $\Phi \in \mathbb{R}^{N\times M}$ is the matrix with elements:
\begin{equation}
    \Phi_{ik}=\psi(\bm{w}_i^{(1)}\bm{y}_k^{(0)}+b_i^{(1)}).
\end{equation}
Therefore, one can find the output weights as a linear least square problem, here using the \texttt{pinv} of \texttt{Matlab}, employing the Moore-Penrose pseudo-inverse (with a regularizing tolerance 1e$-$6).

\paragraph{Random Fourier Features (RFF)} 
\cite{rahimi2007random} is a particularly interesting and straightforward RPNN approach, that uses the cosine as the activation function of the network, i.e., $\psi(\cdot)=cos(\cdot)$. This choice leads to a random Fourier basis expansion approximation. As proposed in \cite{rahimi2007random}, 
we sampled the weights that connect the input and the hidden layer from the following distribution:
\begin{equation}
    p(w)=2\pi^{(-\frac{D}{2})}\text{exp}\biggl(-\frac{||w||^2_2}{2}\biggr),
\end{equation}
where $D$ is the dimension of the input space.
For the choice of the biases we used an uniform distribution in the interval $[0,2\pi]$.

\section{Equation-Free approach}
\label{sec:coarse_timestepper}
The Equation-Free (EF) framework, proposed in \cite{kevrekidis2003equation,kevrekidis2004equation}, operates on the key assumption that for a given microscopic simulator there exists a fundamental coarse description: as the distributions evolve, higher-order moments quickly become dependent on lower-order ones, ultimately converging towards a slow invariant manifold.
This concept embodies the singularly perturbed system paradigm, where the interconnected nonlinear ordinary differential equations governing the moments of the agent distribution rapidly approach a low-dimensional slow manifold. The main tool employed in the EF approach are the so-called \emph{coarse time-steppers}.
Coarse time-steppers establish a link between microscopic simulators, such as ABM and traditional continuum numerical algorithms. Such methods encompass a series of essential stages, as outlined below:
\begin{itemize}
    \item Assume we start from a coarse-scale initial condition, corresponding to the appropriate coarse-scale variables $\bm{x}(t)$ (e.g., diffusion map coordinates) of the evolving ensemble of agents at time $t$;
    \item Map the macroscopic description, $\bm{x}(t)$, through a \textit{lifting operator}, $\mathcal{L}$, to an ensemble of consistent microscopic realizations:
    \begin{equation}
        X(t)=\mathcal{L}(\bm{x}(t));
    \end{equation}
    \item For an (appropriately chosen) short macroscopic time $\Delta T$, evolve these realizations using the black-box ABM simulator to obtain the value(s):
    \begin{equation}
        X(t+\Delta T)=\Phi_{\Delta T}(X(t))\equiv \Phi_{\Delta T}(\mathcal{L}\bm{x}(t));
    \end{equation}
    \item Map the ensemble of agents back to the macroscopic description through the \textit{restriction operator} $\mathcal{M}$:
    \begin{equation}
        \bm{x}(t+\Delta T)=\mathcal{M}X(t+\Delta T).
    \end{equation}
\end{itemize}
The entire procedure, i.e., the coarse time-stepper, can be thought of as a ``black box”:
\begin{equation}
    \bm{x}(t+\Delta T)=\phi_{\Delta T}[\bm{x}(t)]\equiv \mathcal{M}\Phi_{\Delta T}(\mathcal{L}[\bm{x}(t)]).
    \label{eq:EF_black_box}
\end{equation}
Such an approach, allows one to accelerate simulations and also to perform bifurcation analysis, exploiting the Newton-GMRES \cite{kelley1999iterative} algorithm for the computation of equilibria of \eqref{eq:EF_black_box} and Arnoldi iterative algorithms \cite{saad2011numerical} for estimating the dominant eigenvalues of the coarse linearization, quantifying the stability of the steady states.
To find a stationary state $\bm{x}$ of eq.\eqref{eq:EF_black_box}, if there exists, we seek a solution of the following equation:
\begin{equation}
    \bm{F}(\bm{x})=\bm{x}-\phi_{\Delta T}[\bm{x}]=0,
\end{equation}
wrapping around it the Newton-GMRES method.

\end{document}